\documentclass{article}

\usepackage[preprint]{unireps_2024}

\usepackage[utf8]{inputenc} 
\usepackage[T1]{fontenc}    
\usepackage{hyperref}       
\usepackage{url}            
\usepackage{booktabs}       
\usepackage{amsfonts}       
\usepackage{nicefrac}       
\usepackage{microtype}      
\usepackage{xcolor}         

\usepackage{microtype}
\usepackage{graphicx}
\usepackage{subfigure}
\usepackage{booktabs} 
\usepackage{hyperref}

\usepackage{amsmath,amsfonts,bm}

\def\eqref#1{equation~\ref{#1}}

\def\1{\bm{1}}

\DeclareMathAlphabet{\mathsfit}{\encodingdefault}{\sfdefault}{m}{sl}
\SetMathAlphabet{\mathsfit}{bold}{\encodingdefault}{\sfdefault}{bx}{n}

\DeclareMathOperator*{\argmax}{arg\,max}
\DeclareMathOperator*{\argmin}{arg\,min}

\usepackage{xcolor}

\usepackage{amsmath}
\usepackage{amssymb}
\usepackage{mathtools}
\usepackage{amsthm}

\usepackage[capitalize,noabbrev]{cleveref}

\theoremstyle{plain}

\theoremstyle{definition}

\theoremstyle{remark}

\title{Modern Hopfield Networks meet Encoded Neural Representations - Addressing Practical Considerations}

\author{Satyananda Kashyap \\ IBM Research - Almaden 
\\ \texttt{satyananda.kashyap@ibm.com}  
\And Niharika S. D'Souza
\\ IBM Research - Almaden 
\And Luyao Shi 
\\ IBM Research - Almaden 
\And Ken C. L. Wong 
\\ IBM Research - Almaden 
\And Hongzhi Wang 
\\ IBM Research - Almaden 
\And Tanveer Syeda-Mahmood
\\ IBM Research - Almaden 
}

\begin{document}

\maketitle

\begin{abstract}
Content-addressable memories such as Modern Hopfield Networks (MHN) have been studied as mathematical models of auto-association and storage/retrieval in the human declarative memory, yet their practical use for large-scale content storage faces challenges. Chief among them is the occurrence of meta-stable states, particularly when handling large amounts of high dimensional content. This paper introduces Hopfield Encoding Networks (HEN), a framework that integrates encoded neural representations into MHNs to improve pattern separability and reduce meta-stable states. We show that HEN can also be used for retrieval in the context of hetero association of images with natural language queries, thus removing the limitation of requiring access to partial content in the same domain. Experimental results demonstrate substantial reduction in meta-stable states and increased storage capacity while still enabling perfect recall of a significantly larger number of inputs advancing the practical utility of associative memory networks for real-world tasks. 
\end{abstract}  

\section{Introduction}

The hippocampal system in the brain plays a central role in long-term memory, responsible for storing and recalling facts and events. Its structure, particularly the auto-associative networks in the CA3 region, enables efficient memory retrieval based on partial input, a process that has inspired several mathematical models of memory networks \citep{coldspringmemory,whittington2020tolman, gillett2020characteristics, burns2022multiscale}. Classical Hopfield network models~\citep{hopfield1982neural, amari1972learning} are a type of associative memory architecture that stores memories as fixed point attractor states in an energy landscape, using Hebbian learning to recall patterns form partial input cues using a recurrent network. The Modern Hopfield network (MHN) introduced a continuous relaxation of the original method and in theory enable exponential storage capacity growth with respect to the number of neurons~\citep{krotov2016dense, demircigil2017model, ramsauer2020hopfield}. MHNs have been applied to tasks such as immune repertoire classification~\citep{vaswani2017attention, widrich2020modern} and graph anomaly detection~\citep{hoover2023energy}.

Extending these ideas, heteroassociative memories provide a framework for hippocampal memory storage and retrieval, enabling pattern recall from different input modalities. Hetero-association has also been interpreted in various ways, including for modeling sequence associations~\citep{chaudhry2023long, karuvally2023general, gutfreund1988processing}, where the process begins with a given pattern in a sequence, adjusting the energy weights to transition from one pattern to the next. Similarly, \citep{tyulmankov2021biological} introduces a key-value memory model where sequential patterns are used to predict the next item in the sequence. Further, \citep{millidge2022universal} demonstrated hetero-association by reconstructing missing portions of an image based on other parts of the same image. Most research on content addressable memories focus on dense associative memory theory, and use simplistic scenarios or simulations for validation~\citep{kang2023hopfield, iatropoulos2022kernel}. Translating to practical large-scale content storage systems has been difficult, as associative memories like MHNs tend to enter spurious meta-stable states when handling large volumes of content~\citep{martins2023sparse}. The metastable states are closely related to the separability of the input patterns~\citep{ramsauer2020hopfield}. Additionally, MHNs rely on partial content for recall, which limits their utility in cases where such cues are unavailable.

In this paper we address the issue of meta-stable states in Modern Hopfield Networks (MHN) by improving the separability of input patterns through neural encoding representations. Our approach termed HEN (modern Hopfield networks with Encoded Neural representations or in short Hopfield Encoding Networks) encodes inputs into a latent representational space using a pre-trained neural encoder-decoder model before storage, and decodes them upon recall. While prior work has explored encoding content for associative memory ~\citep{kang2023hopfield}, our method uniquely combines pre-encoding and post-decoding to specifically tackle metastable states in MHNs. Additionally, HEN supports hetero-association, allowing retrieval through free text queries, thus eliminating the need for users to provide partial content for recall. Comprehensive ablation experiments demonstrate that HEN significantly increases storage capacity, reduces metastable states across modalities, and enables perfect recall of a significantly larger number of stored elements using natural language input.

Our approach begins by surveying different representations of energy-based formulations, demonstrating how they form a unifying framework that connects support vector-based kernel memories~\citep{iatropoulos2022kernel}, Modern Hopfield networks~\citep{ramsauer2020hopfield, krotov2020large}, and the transformer attention mechanism~\citep{vaswani2017attention}. By establishing these theoretical connections, we shed light on how similar representations emerge in distinct neural models.

Despite the theoretical promise of MHNs, we identify their practical limitations, notably the emergence of spurious states due to weak input pattern separability. \textbf{To address this issue, we propose HEN, which enhance pattern separation by encoding input representations into a latent space before storage.} This method delays the onset of metastability and significantly increases storage capacity, leading to improved recall stability compared to alternative representation learning strategies.

\section{Modern Hopfield Networks: A Representational Perspective}
\label{sec:preliminaries}

In this section, we review the basic framework of Modern Hopfield Networks (MHNs), focusing on their representational properties, equivalence with other models, and inherent limitations. This discussion motivates our enhancements to MHNs using neural encoding strategies aimed at improving pattern separability and storage capacity.

The MHN framework provides a framework for dense associative memory using continuous dynamics. It can be described in terms of its energy function and the resulting attractor dynamics. Let $N$ be the number of memories and $K$ be the data dimensionality (number of neurons in the MHN). Defining a similarity metric between the memories $\{\mathbf{\xi}_{n}  \in \mathcal{R}^{K \times 1} \}^{N}_{n=1} $ or matrix $\mathbf{\Xi} \in \mathcal{R}^{N\times K}$ and the state vector $\mathbf{s} \in \mathcal{R}^{K \times 1}$ the generalized objective function is expressed as the following energy minimization: 
\begin{gather}
    \mathbf{s}^{*} = \argmin_{\mathbf{s}} E(\mathbf{s},\mathbf{\Xi}) = \argmin_{\mathbf{s}} { E_{1}(\mathbf{s},\mathbf{\Xi};{\beta}) + E_{2}(\mathbf{s})} \label{Energy}  \\
    E_{1} (\mathbf{s},\mathbf{\Xi};{\beta})
    =  F_{\beta} (f_\text{sim}(\{\mathbf{\xi}_n\},\mathbf{s})) =  F_{\beta}\Big(\{\sum_{k=1}^{K} \mathbf{\xi}_{n}(k) \mathbf{s}(k)\}\Big) = -\frac{1}{\beta}\log{\Big[\sum^{N}_{n=1}\exp\Big( \beta \sum_{k=1}^{K} \mathbf{\xi}(k) \mathbf{s}(k)\Big)\Big]} \nonumber \\
    E_{2}(\mathbf{s}) = \frac{1}{2}\mathbf{s}^{T}\mathbf{s} + \text{constant} \nonumber
\end{gather}

The function $f_{\text{sim}}(\{\mathbf{\xi}_{n}\}, \mathbf{s}; \beta)$ is a measure of similarity between the state vector and each memory in the bank. A common choice for this similarity metric is the dot product in $K$ dimensional vector space, which is both efficient and widely adapted in practical implementations of MHNs ~\citep{ramsauer2020hopfield, krotov2020large}. The energy function $F_{\beta}(\cdot)$, defined as the log-sum-exponential (LSE), approximates the $\argmax$ function to select the most relevant memory. The inverse temperature parameter $\beta$ controls the sharpness of this selection. The energy minimization equation treats memories as attractors of a dynamical system~\citep{krotov2016dense}, recoverable from partial cues through an iterative optimization process. The state vector recurrence is given by: 

\begin{equation}
    \mathbf{s}^{(t+1)} = \mathbf{\Xi}  \text{softmax}(\beta  \mathbf{\Xi}^{T}   \mathbf{s}^{(t)}) = \frac{\sum_{n} \mathbf{\xi}_{n} \exp(\beta \xi^{T}_{n} \mathbf{s}^{(t)})}{\sum_{n}\exp(\beta \xi^{T}_{n} \mathbf{s}^{(t)})}
    \label{dynamics}
\end{equation}

These updates progressively reduce the energy of the system monotonically \citep{millidge2022universal} and are guaranteed to converge under specific conditions \citep{ramsauer2020hopfield}. Starting with an initial state $\mathbf{s}^{(0)}$, which could be a partial or noisy memory cue, the iterations  $\{\mathbf{s}^{(t)}\}$ aims to reconstruct a full pattern $\mathbf{s}^{(T_{f})}$ that corresponds to one of the stored memories $\{\mathbf{\xi}_{n}\}$. However, in practice, the process may lead to local minima or saddle points, resulting in meta-stable configurations.

\subsection{Equivalence with Kernel Memory Networks (KMN)}
\label{equivalence}

The formulation of the MHNs can be viewed as a special case of Kernel Memory Networks (KMNs)~\citep{iatropoulos2022kernel}, where each neuron performs kernel-based classification or regression. The MHN update rule is analogous to that in KMNs and can be computed in closed form as a recurrence. 
Let ${K}(\mathbf{\Xi}, \mathbf{s}^{(t)})$ denote the (symmetric positive definite) kernel function that defines pairwise similarities, and $\mathbf{K}^{\dagger}$ be the Moore-Penrose pseudoinverse for $\mathbf{K}(\mathbf{\Xi},\mathbf{\Xi})$. For continuous valued memories, the radial translation-invariant exponential kernel (infinite dimensional basis) with a fixed spatial scale $r$ and temperature $\alpha$ is proposed. This parameterization allows for an analysis of memory capacity and storage limits by interpreting the MHN optimization as a feature transformation operating in a Reproducing Kernel Hilbert Space (RKHS). The kernel and state update are as follows:

\begin{equation}
    \text{Kernel:} \ \ K_{(\alpha,r)}(\mathbf{x},\mathbf{y}) = \exp{\Big[ - \Big(\frac{1}{r} {\vert\vert{\mathbf{x}-\mathbf{y}}\vert\vert}_{2}\Big)^{\alpha}\Big]} \ \ \ \ \text{State Update:} \ \  \mathbf{s}^{(t+1)} = \mathbf{\Xi} \mathbf{K}^{\dagger} {K}(\mathbf{\Xi}, \mathbf{s}^{(t)})
    \label{kernelmemory}
\end{equation}
While KMNs provide strong theoretical storage guarantees~\citep{iatropoulos2022kernel}, their real-world performance is known to be sensitive to data distributions and parameter choices~\cite{wu2024uniform}.
\subsection{Equivalence with Transformers}
\label{transformers}
Alternatively, the update rule in Eq.~(\ref{dynamics}) has been shown~\citep{ramsauer2020hopfield} to be equivalent to the key-query \textit{self}-attentional framework used in transformer models~\citep{vaswani2017attention}.
\begin{equation}
\mathbf{Z} = \text{softmax}\Big(\frac{1}{\sqrt{d_{k}}}\mathbf{Q}\mathbf{K}^{T}\Big)\mathbf{V} \quad \text{or} \quad \mathbf{Z}^{T} = \textbf{V}^{T} \text{softmax}\Big(\frac{1}{\sqrt{d_{k}}}\mathbf{K}\mathbf{Q}^{T}\Big)
\label{transformer}
\end{equation}
with the keys being related to the memories as $\mathbf{W}_{K}\mathbf{\Xi} = \mathbf{K}$, and queries/values being related to the intermediate state vectors $\mathbf{s}^{t+1} = \mathbf{Z}^{T}$, $\mathbf{W}_{Q}\mathbf{s}^{t} = \mathbf{Q}^{T}$, $\mathbf{W}_{V}\mathbf{s}^{t} = \mathbf{V}^{T}$ and the dispersion parameter relating to the temperature $\frac{1}{\sqrt{d_{k}}} = \beta$. The matrices $\mathbf{W}_{Q}, \mathbf{W}_{K}, \mathbf{W}_{V} :\mathcal{R}^{K} \rightarrow \mathcal{R}^{D} $ are linear transformations associated with the query-key-value triplet, which when substituted with the identity matrix $\mathcal{I}_{K}$ gives us the form in Eq.~(\ref{dynamics})

Under the kernel memory networks framework, the MHN equations (\ref{Energy}-\ref{dynamics},\ref{transformer}) do not involve a symmetric positive definite kernel, as the energy objective is inherently non-convex~\citep{iatropoulos2022kernel,wright2021transformers}. However, the system still permits a bilinear reproducing form for the kernel $K(\cdot,\cdot)$, where input patterns are mapped into higher-dimensional feature spaces. This bilinear kernel is equivalent to the transformer's key-query attention mechanism, where inputs are projected into higher dimensional feature spaces.

The transformer kernel can be written as: 
$K(\mathbf{x},\mathbf{y}) = \exp{\Big[\frac{1}{\sqrt{d_{k}}} (\mathbf{W}_{Q}\mathbf{x})^T(\mathbf{W}_K\mathbf{y})\Big]}$.
Overall, this highlights the connection between the MHNs and transformer attention mechanisms, showing that both rely on projecting input representations into a shared space for similarity-based comparison. Further details on the KMN and transformer equivalence are provided in {Appendix Section 1}.
\section{HEN: Modern \underline{H}opfield Networks with \underline{E}ncoded \underline{N}eural Representations}
\label{Encoded}

Although the results of modern Hopfield networks and its various equivalent forms imply that the formulation has theoretically exponential capacity to store memories, \textit{our experimental results demonstrate that the system of updates can be brittle in practice and highly sensitive to real-world data distributions across each of these data representations}. 

Specifically, MHN often struggle with spurious attractor basins~\citep{bruck1990number, ramsauer2020hopfield, barra2018new}, which manifest as erroneous memory patterns due to overlapping or similar inputs. This issue is particularly evident in large datasets, where poor pattern separability results in meta-stable states, limiting retrieval accuracy and scalability. 

The key insight from the unified representations discussed in Sections ~\ref{equivalence} and~\ref{transformers} is that improving the separability of input memories significantly enhances retrieval accuracy. This can be achieved by mapping input memories $\mathbf{\Xi}$ and partial queries $\mathbf{s}^{(0)}$ from their original $K$-dimensional space (where memories may overlap or be less distinct) into a higher dimensional embedding space, the stored patterns become better separable. The increased separability directly reduces the occurrence of spurious attractors and leads to more reliable memory retrieval. 

Going one step further, a key idea we put forward here is to see if we can bolster the separability of the input patterns before they enter the Modern Hopfield network in order to reduce the spurious attractor states problem via large pre-trained encoded-decoder models and their latent space representations (i.e. generalizing the linear transforms in Eq.~\ref{transformer}). Following observations of the phenomenon of input encoding in the dentate gyrus (DG) region of the trisynaptic circuit prior to memorization~\citep{dghopfield}, we propose to store these latent-space neural encodings, i.e. transformation computed on the memories $\hat{\mathbf{\Xi}}$ and partial query $\hat{\mathbf{s}}^{(0)}$, in the MHN memory bank using the encoder transformation $\mathbf{\Phi}_{\text{enc}}(\cdot)$. Recovery of such patterns can be performed by unrolling the recurrence relation in the latent-space (i.e. Eq.(\ref{dynamics})) followed by applying the associated decoder transformation $\mathbf{\Phi}_{\text{dec}}(\cdot)$ to the latent space representation $\hat{\mathbf{s}}^{(T_{f})}$. Mathematically, this procedure can be expressed as follows:
\begin{gather}
\hat{\mathbf{\Xi}} = \mathbf{\Phi}_{\text{enc}}(\mathbf{\Xi}) \ \ \ \hat{\mathbf{s}}^{(0)} = \mathbf{\Phi}_{\text{enc}}(\mathbf{s}^{(0)}) \\
    \hat{\mathbf{s}}^{(t+1)} = \hat{\mathbf{\Xi}}  \text{softmax}(\beta  \hat{\mathbf{\Xi}}^{T}   \hat{\mathbf{s}}^{(t)}) = \frac{\sum_{n} \hat{\mathbf{\xi}}_{n} \exp(\beta \hat{\xi}^{T}_{n} \hat{\mathbf{s}}^{(t)})}{\sum_{n}\exp(\beta \hat{\xi}^{T}_{n} \hat{\mathbf{s}}^{(t)})}
\\
\mathbf{s}^{(T_{f})} = \mathbf{\Phi}_{\text{dec}}(\hat{\mathbf{s}}^{(T_{f})})
\label{HEN}
\end{gather}

{\em Hypothesis 1: The spurious attractors can be reduced by encoding inputs prior to storing them in the Modern Hopfield network and decoding them after recall due to increased separability in latent space}

Our proposed HEN combines an auto-encoder with the Modern Hopfield network (MHN). Specifically, the encodings are generated by a pre-trained auto-encoder. The raw content is then recovered through chaining MHN with the decoder portion of the auto-encoder. We hypothesize that the encodings produced by an auto-encoder contain discriminative information that is not only compact but can improve the separability in the energy landscape to significantly delay the meta-stable states even with increased content. That is, by leveraging a well-trained auto-encoder for feature extraction, we posit that the most significant and discernible features between images can be easily identified, leading to less spurious patterns emerging during recall and allowing more content to be stored, thereby increasing storage capacity.
\begin{figure*}[b]
    \centering
\includegraphics[width=0.8\textwidth]{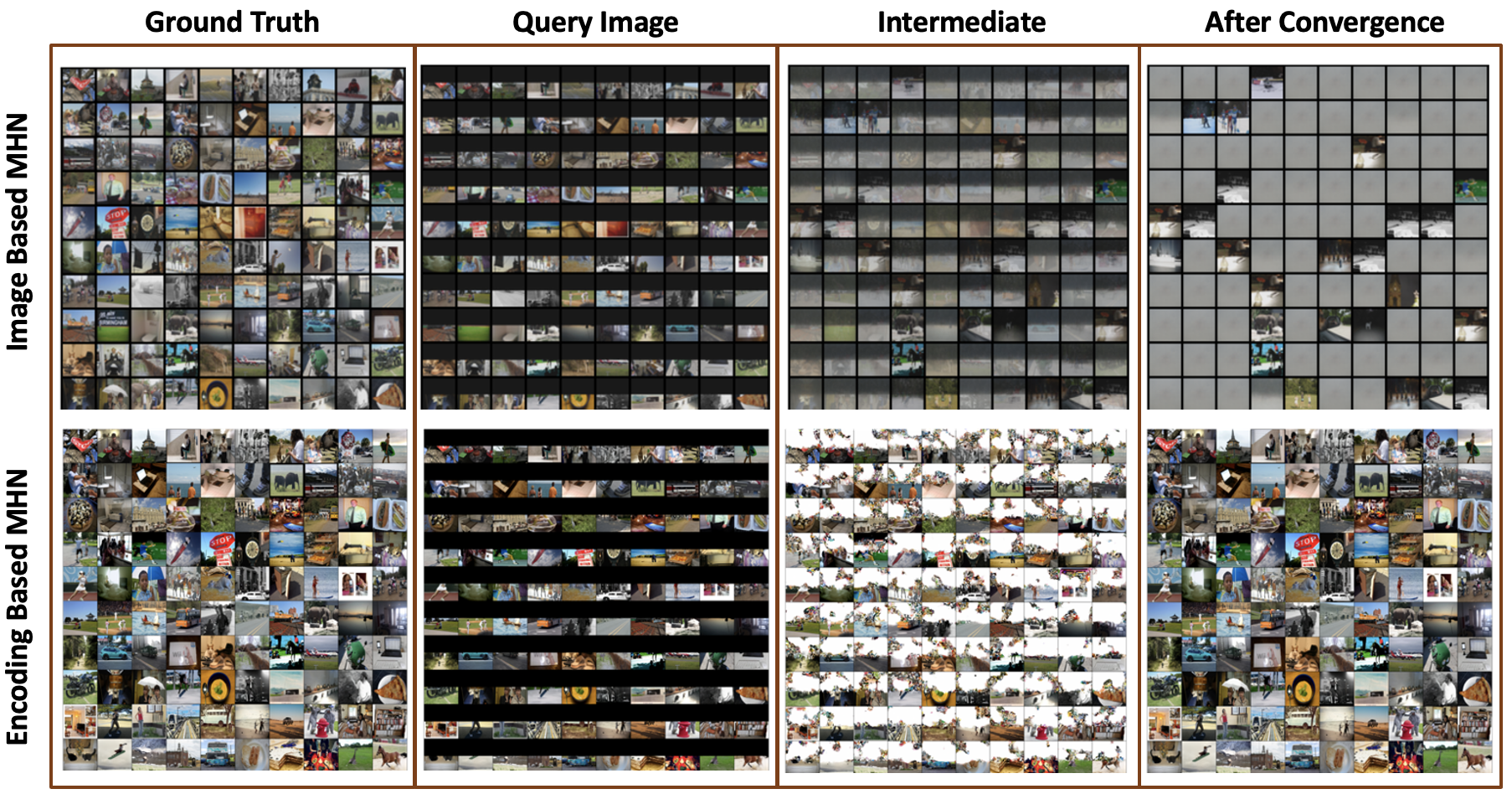}
\caption{Progression of the \textbf{(Top Row)} Modern Hopfield Network (MHN) run on image inputs and the \textbf{(Bottom Row)} HEN at different intermediate steps. The sequence from left to right is as follows: the original image, the query image with half of it occluded, an intermediate update at iteration 11, and the final reconstruction at iteration 150. This uses an $\ell_2$ similarity and discrete Variational Autoencoder (D-VAE) encoder for the HEN. We set $\beta=150$ in both experiments}
\label{fig:sparsity_image}
\end{figure*}
\begin{figure*}[b]
\centering
\includegraphics[width=0.8\textwidth]{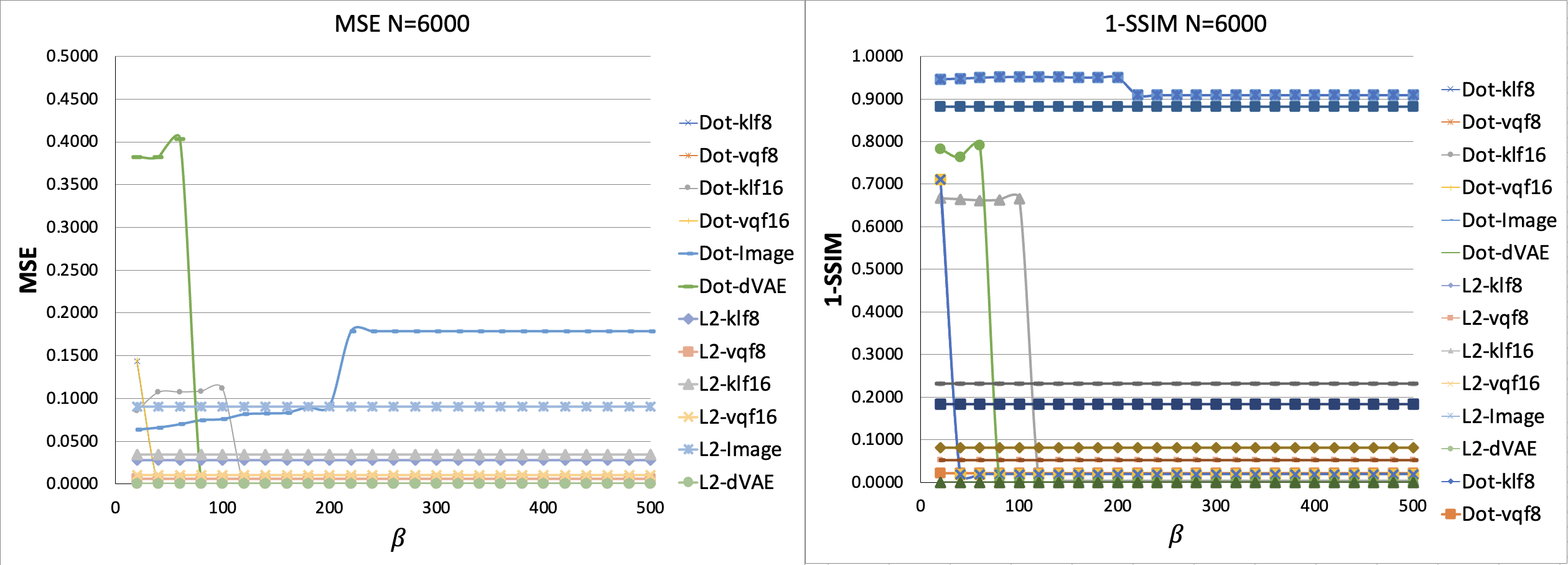}
\caption{Memory recall performance of various encoder methods and the image-based Modern Hopfield network, each color-coded differently. The \textbf{(Left)} figure plots the MSE while the \textbf{(Right)} depicts the 1-SSIM as a function of $\beta$. The encoder-based HEN methods outperform the raw image-based method over a very large range of choices of hyperparameters. Dot or L2 in the legend denote the dot product or $\ell_2$, followed by the type of representation used including Original input-Image, dVAE-Discrete Variational Auto Encoder, Kullback-Leibler (KL)-based variants- KLf8, KLf16, and Vector quantized VAE methods-VQf8, VQf16 per the convention in ~\citep{rombach2021high}}
    \label{fig:beta-ablation}
\end{figure*}

\subsection{Experimental Evaluation and Results:} 

We provide practical insights drawn from experiments on the MS-COCO dataset, which contains 110,000 images~\citep{lin2015microsoft}. This dataset offers a more realistic distribution of high-dimensional, real-world data compared to smaller, curated datasets like MNIST. Its unique associative captions also make it ideal for illustrating hetero-associations within our proposed framework.

To evaluate this hypothesis, we conducted studies that examined the effectiveness of various pre-trained encoder-decoder architectures to produce encoded representations that can lead to successful recall of dense associative memories by comparing them against the native data representations and KMNs. We also analyzed the parameter choices for the energy formulation of MHNs in affecting the identity of the recalled memory items when using their encoded representations.

Specifically, we evaluated various pre-trained encoder-decoder architectures known for their state-of-the-art performance in deep learning-based image encoding and decoding ranging from vanilla-transformer based to variational auto-encoder based models. In particular, we utilized the Discrete Variational Autoencoder (D-VAE) from~\citep{ramesh2021zero} and other architectures from~\citep{rombach2021high} and explored two transformer variants from the diffusion library: one trained with codebook-based (Vector Quantized - VQ) criteria and the other using Kullback-Leibler (KL) divergence-based criteria. Our empirical analysis revealed that Vector Quantized VAE (VQ-VAE) methods outperformed others in our setup. Consequently, we selected D-VAE and variants of VQ from \citep{rombach2021high} for further analysis. To maintain consistency in representation, we downsampled all the images to a resolution of \(28 \times 28 \times 3\) to match the number of features that the encoded representations produced.

\noindent{\textbf{Evaluation Metrics}}: This study tested the image-based dense MHNs (Eq.~\ref{Energy}) and KMNs with the exponential kernel against pre-trained Discrete VAE~\citep{ramesh2021zero} and VQ-VAEs encoding equipped Hopfield encoding network \citep{rombach2021high} (Eq.~\ref{HEN}). The test was conducted on a memory bank ($\{\mathbf{\xi}_{n} \in \mathcal{R}^{1 \times K}\}$) and query size ($N$) of 6000 images from the MS-COCO dataset. To examine the effect of different choices of $f_{\text{sim}}(\cdot,\cdot)$ in Eq.~(\ref{Energy}), both dot product and $\ell_2$ based similarity measures were utilized. The performance of different encoder-decoder architectures was evaluated (see Fig.~\ref{fig:beta-ablation}) by varying the dimensionality $K$. The Mean Squared Error ($MSE$) and Structural Similarity Index ($1-SSIM$) metrics were used to compute the similarities between the encoder reconstructions stored in the $\{\mathbf{\xi}_{n}\}$ memory bank and the reconstructed ones.

Note that as this evaluation was conducted to assess the performance on metastable states, \textit{the focus was on recovering the correct identity rather than the quality of reconstruction}. Hence, a $MSE=1-SSIM=0$ indicated that the dense associative memory could retrieve the full encoded representation of the image from which the pre-trained decoder could reconstruct the image. \footnote{We run a sanity check based on the column rank of the recovered pattern matrix $\hat{\mathbf{S}}^{(T_{f})}$ to illustrate that these metrics correlate well with degeneracy in the recovered patterns. See Section 3 in the Appendix.}

\noindent\textbf{Results:} Figure~\ref{fig:beta-ablation} shows the result of our analysis using six different encoding methods including raw image store and two similarity types (dot product and $\ell_2$). The encoded representations uniformly perform well above a certain $\beta$ value. In comparison, we note that the baseline image-based MHN (Dot-image \& L2-image) persists in meta-stable states irrespective of the $\beta$ value or the similarity metric. Additionally, we also assessed metastable states by recording the relative reduction in the rank of the update matrix with the collapse of the pattern recovery process. The result is available in the Appendix, Section 2 and shows that for a judiciously chosen value of $\beta$, the iterates of HEN stabilize to provide near perfect retrieval, consistent with the trends observed in Figure~\ref{fig:beta-ablation}. 

\noindent{\textbf{KMNs Sensitivity to Real-World Conditions}}:
KMNs have strong theoretical guarantees, but in our experiments, they faced significant challenges with real-world, large-scale image retrieval tasks. Despite an extensive parameter sweep over hyperparameters like $r$ (spatial scale) and $\alpha$ (inverse temperature), KMNs were highly sensitive to these settings and consistently underperformed. For example, with a memory bank of 6000 images, KMNs yielded an MSE of 0.2427 and 1-SSIM of 0.972, indicating poor retrieval accuracy. While KMNs excel in controlled environments, their performance degrades in high-dimensional, non-linearly separable data, limiting their practical use. This is suspected to be a direct consequence of the restrictive assumptions made on the underlying data-distributions in KMNs~\citep{wu2024uniform}. In contrast, our HEN shows more robust performance, mitigating meta-stable states and improving retrieval accuracy.

\noindent{\bf Quality of recovery}: Fig.~\ref{fig:sparsity_image} shows the result of using  D-VAE encoding for perfect memory recall for the same set of images for which raw image storage in the Hopfield network failed. While the reconstruction quality is not as clear as the original, the identity of the recovered images is preserved one-to-one. In comparison, the recall using the raw images for the same dataset using the Modern Hopfield network shows the metastable states.

\noindent{\bf Scale-out Performance and Ablation Study on Encoder-Decoder Pairs}: We evaluated HEN's scalability by testing it's retrieval performance as the number of stored images increased in the memory bank. Table \ref{tab:half-image-results} presents the performance metrics across different encoder-decoder approaches as the memory bank scaled from 6,000 to 15,000 images. Our results show that all the encoder-based approaches robustly recovered the image representations without a noticeable drop in the reconstruction quality, supporting HEN's stability and scalability in  retrieval performance at this scale. Additionally, our ablation studies included five pre-trained encoder-decoder architectures - dVAE ~\citep{ramesh2021zero} and  VQ-F8, VQ-F16, KL-F8, KL-F16 from \citep{rombach2021high} to assess whether HEN's retrieval stability holds across varied configurations. Section 2 of the Appendix contain a more detailed breakdown of the recovery performance as a function of encoder-decoder methods, and the similarity metric for different values of $\beta$. \textit{{By contrast, KMNs consistently failed to scale with the increase in images, showing degraded performance as the number of images increased.}}

\begin{table}[t]
\caption{All neural encoders can recover images without quality loss as more data is stored in $\{\mathbf{\xi}_{n}\}$.}
\centering
\medskip
\resizebox{0.6\columnwidth}{!}{%
\begin{tabular}{@{}ccccc@{}}
\toprule
\multicolumn{1}{l}{\begin{tabular}[c]{@{}l@{}}NUM IMAGES\\ $1-SSIM$, $MSE$\end{tabular}} & \multicolumn{1}{l}{vqf8} & \multicolumn{1}{l}{vqf16} & \multicolumn{1}{l}{Image} & \multicolumn{1}{l}{D-VAE} \\ \midrule
6000 & 0.021, 0.000 & 0.019, 0.004 & 0.836, 0.064 & \textbf{0.000, 0.000} \\ \midrule
8000 & 0.019, 0.000 & 0.046, 0.004 & 0.835, 0.067 & \textbf{0.000, 0.000} \\ \midrule
10000 & 0.019, 0.000 & 0.047, 0.004 & 0.835, 0.064 & \textbf{0.000, 0.000} \\ \midrule
15000 & 0.019, 0.000 & 0.048, 0.004 & 0.836, 0.066 & \textbf{0.000, 0.000} \\ \bottomrule
\end{tabular}%
\label{tab:half-image-results}}
\end{table}

\begin{figure*}[t]
\centering
\includegraphics[width=0.8\textwidth]{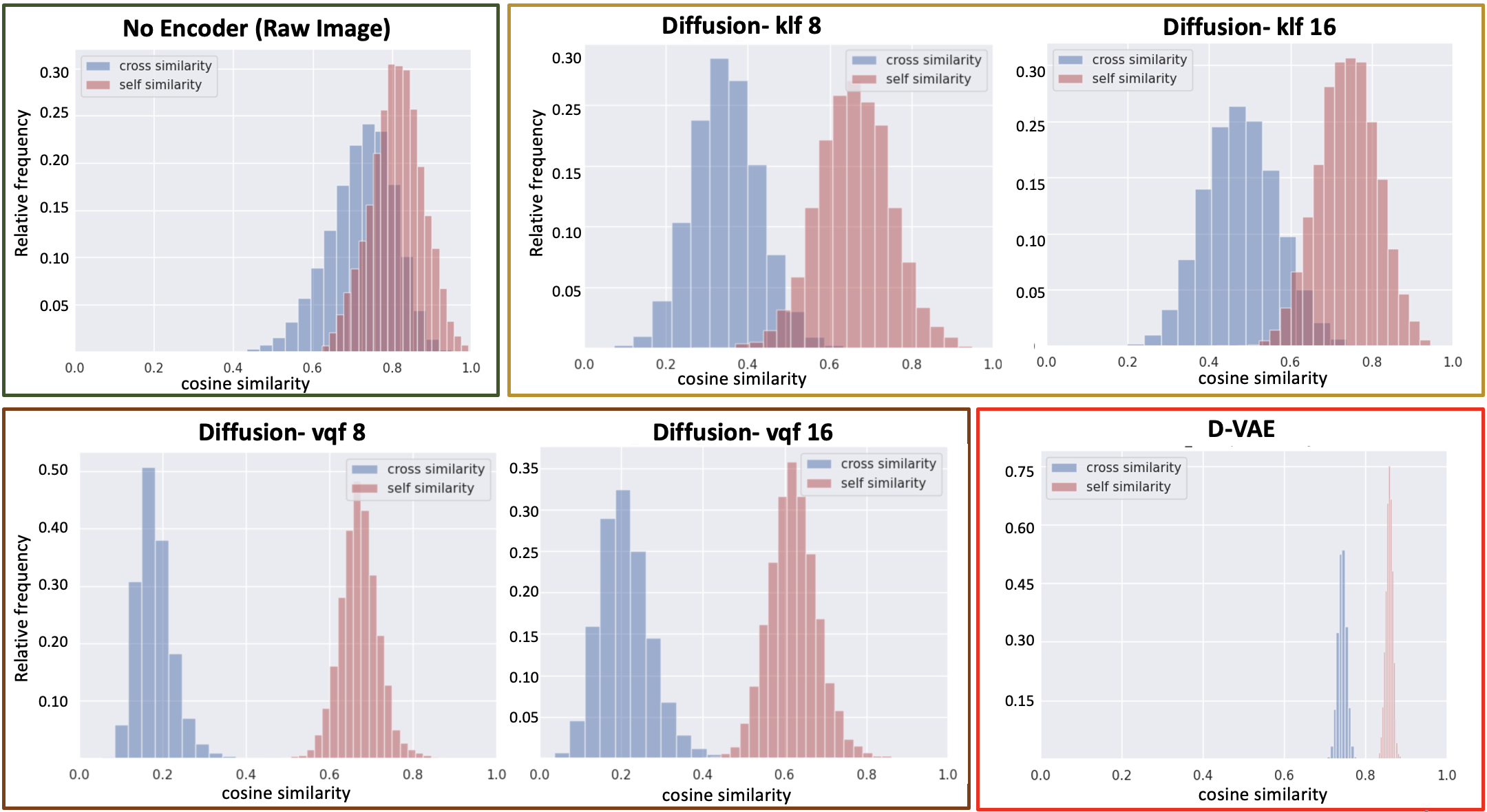}
\caption{Illustrating separability in various embeddings used in Subsection~\ref{Encoded}. We plot the histogram of the distribution of cosine similarity values between the queries and memories, i.e. $\text{cos}(\hat{\mathbf{S}}_{i}^{(0)},\hat{\mathbf{\xi}}_{j})$. Distributions colored indicate self-similarity (i.e. $i=j$) across paired examples, while distributions colored in blue indicate cross similarities (i.e. $i \neq j$). We generate these distributions for (a) Raw Images in the Black Box,  Diffusion models trained (b) on KL Divergence in the Orange Box, (c) trained using Vector Quantization in the Brown Box, and (d) Discrete-VAE (D-VAE) in the Red Box.}
\label{fig:visual_study}
\end{figure*}

\noindent \textbf{Probing the separability of HEN encodings:} To study separability of various encoding strategies,  we examine the strength of association patterns in the HEN memory bank, i.e. the latent-space vectors. Extending the notation in Section~\ref{Encoded}, let $\hat{\mathbf{S}}_{i}^{(0)} \in \mathcal{R}^{K \times 1} = \mathbf{\Phi}_{\text{enc}}(\mathbf{S}_{i}^{(0)})$ denote the encoded query for example $i$ in the dataset. This encoding is generated by occluding a portion of the image fed through the encoder (or just the occluded image for the raw image MHN). We expect that the major contributor to poor recovery performance is the lack of separation between the attractor basins in in Eq.~(\ref{Energy}), due to which the dynamics of state evolution $\hat{\mathbf{S}}^{(t)}_{i}$ in Eq.~(\ref{dynamics}) are meta-stable configurations. To quantify this separation, we compute the cosine similarity between pairs of query and memory vectors, i.e. $c_{ij} = \text{cos}(\hat{\mathbf{S}}_{i}^{(0)},\hat{\mathbf{\xi}}_{j}) = {\hat{\mathbf{\xi}}^{T}_{j}}\hat{\mathbf{S}}_{i}^{(0)}/{{\vert\vert{\hat{\mathbf{S}}_{i}^{(0)}}\vert\vert}_{2}{\vert\vert{\hat{\mathbf{\xi}}_{j}}\vert\vert}}_{2}$. If the patterns are well separated, each query $\hat{\mathbf{S}}_{i}$ in the encoding space (or in the native space for raw images) is close to its own memory $\hat{\mathbf{\xi}}_{i}$ but far apart from others $\hat{\mathbf{\xi}}_{j}, \forall j \neq i$. We test this in Fig.~\ref{fig:visual_study} by plotting the distribution of values as histograms for $c_{ij} \forall j\neq i$ colored in blue and for $c_{ii}$ colored in red, for different $\mathbf{\Phi}_{\text{enc}}({\cdot})$s in HEN. 

This separation is poor for the raw image case, with the two histograms having a high overlap in values. This overlap substantially reduces across the neural encoder-based models, with the Vector Quantized variants providing improved separability and a relatively higher magnitude of self-similarity values compared to their KL counterparts. Finally, we notice that the D-VAE encoder, besides providing separable encodings, also results in the tightest fit around the mean for the self and cross-similarity value distributions. This is likely why the D-VAE provided the best performance (Fig.~\ref{fig:beta-ablation} and Table~\ref{tab:half-image-results}).

\subsection{Natural language-based  Hetero-associations}
\label{NLcs}
We now extend the HEN framework in a hetero-associative setting as a practical application. Specifically, we explore the use of cross-stimuli coming from language and vision, as language-based queries are often used as cues for recall, for example, in practical storage/retrieval contexts. Cross-associative features have been previously demonstrated for the classical Hopfield networks model, albeit under the limited setting of carefully curated binary patterns~\citep{shriwas2019multi}.

{\em Hypothesis 2: Hopfield encoding networks serve as content-addressable memories even with cross-stimuli associations as long as they are unique associations.}

\begin{figure*}[t!]
\centering   \includegraphics[width=0.7\textwidth]{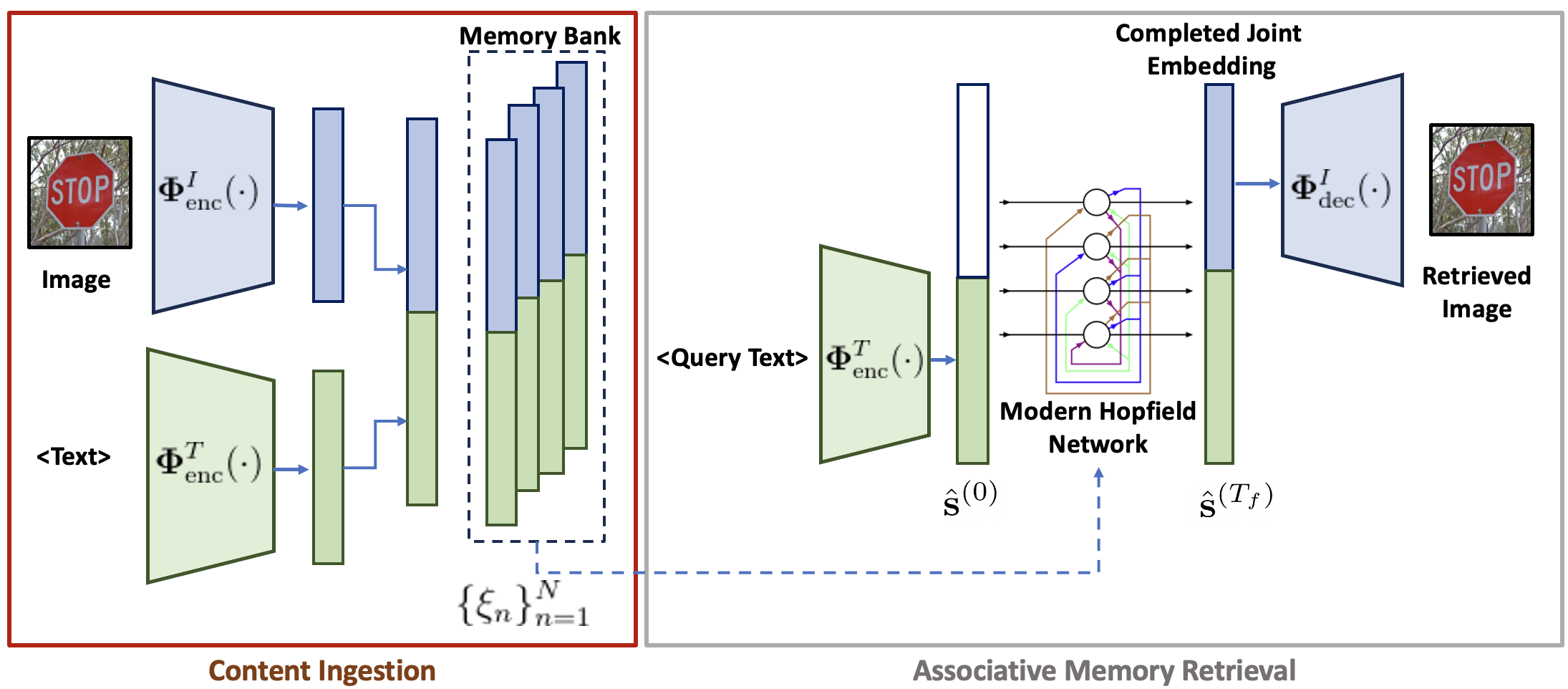}
\caption{HEN architecture for Natural language based hetero-associations. \textbf{Orange Box:} Paired text and image inputs are fed to text ($\mathbf{\Phi}^{\mathbf{T}}_{\text{enc}}(\cdot)$) and image ($\mathbf{\Phi}^{\mathbf{I}}_{\text{enc}}(\cdot)$) encoders respectively to generate the memories of the HEN memory bank. \textbf{Grey Box:} At query time, a partial query $\hat{\mathbf{s}}^{(0)}$ is generated by feeding the query text into $\mathbf{\Phi}^{\mathbf{T}}_{\text{enc}}(\cdot)$. After convergence, the image encoding is extracted from $\hat{\mathbf{s}}^{(T_{f})}$, and decoded through the Image decoder ($\mathbf{\Phi}^{\mathbf{I}}_{\text{dec}}(\cdot)$) to retrieve the corresponding image. Using full image representations instead of image encodings implies $\mathbf{\Phi}^{\mathbf{I}}_{\text{enc}}(\cdot) = \mathbf{\Phi}^{\mathbf{I}}_{\text{dec}}(\cdot) = \mathcal{I}_{K}$, the identity transformation. In the experiment where the text captions are pixelized as input, $\mathbf{\Phi}^{\mathbf{T}}_{\text{enc}}(\cdot) = \mathbf{\Phi}^{\mathbf{I}}_{\text{enc}}(\cdot)$. }
	\label{fig:MHN}
\end{figure*}
\begin{table*}[h!]
\caption{Performance of encoded cross-modal HEN compared to image-based MHNs as the memory bank $\{\hat{\mathbf{\xi}}_{n}\}$ increases. The first row shows the CLIP-encoded cross-modal representations, while the rest present pixelized text-encoded representations for increasing dataset sizes.}
\label{tab:my-table}
\centering
\medskip
\resizebox{0.6\columnwidth}{!}{%
\begin{tabular}{@{}ccccc@{}}
\toprule
\multicolumn{1}{l}{\begin{tabular}[c]{@{}l@{}}NUM IMAGES\\ 1-SSIM, MSE\end{tabular}} & \multicolumn{1}{l}{vqf8} & \multicolumn{1}{l}{vqf16} & \multicolumn{1}{l}{Image} & \multicolumn{1}{l}{D-VAE} \\ \midrule
6000-CLIP & 0.016, 0.000 & 0.024, 0.000 & 0.681, 0.118 & \textbf{0.000, 0.000} \\ \midrule
6000 & 0.016, 0.000 & 0.024, 0.000 & 0.952, 0.214 & \textbf{0.000, 0.000} \\ \midrule
8000 & 0.016, 0.000 & 0.023, 0.000 & 0.952, 0.215 & \textbf{0.000, 0.000} \\ \midrule
10000 & 0.015, 0.000 & 0.024, 0.000 & 0.952, 0.215 & \textbf{0.000, 0.000} \\ \midrule
15000 & 0.015, 0.000 & 0.024, 0.000 & 0.952, 0.215 & \textbf{0.000, 0.000} \\ \bottomrule
\end{tabular}%
\label{tab:cross-modal-results}
}
\end{table*}
\begin{figure*}[b!]
\centering    \includegraphics[width=0.8\textwidth]{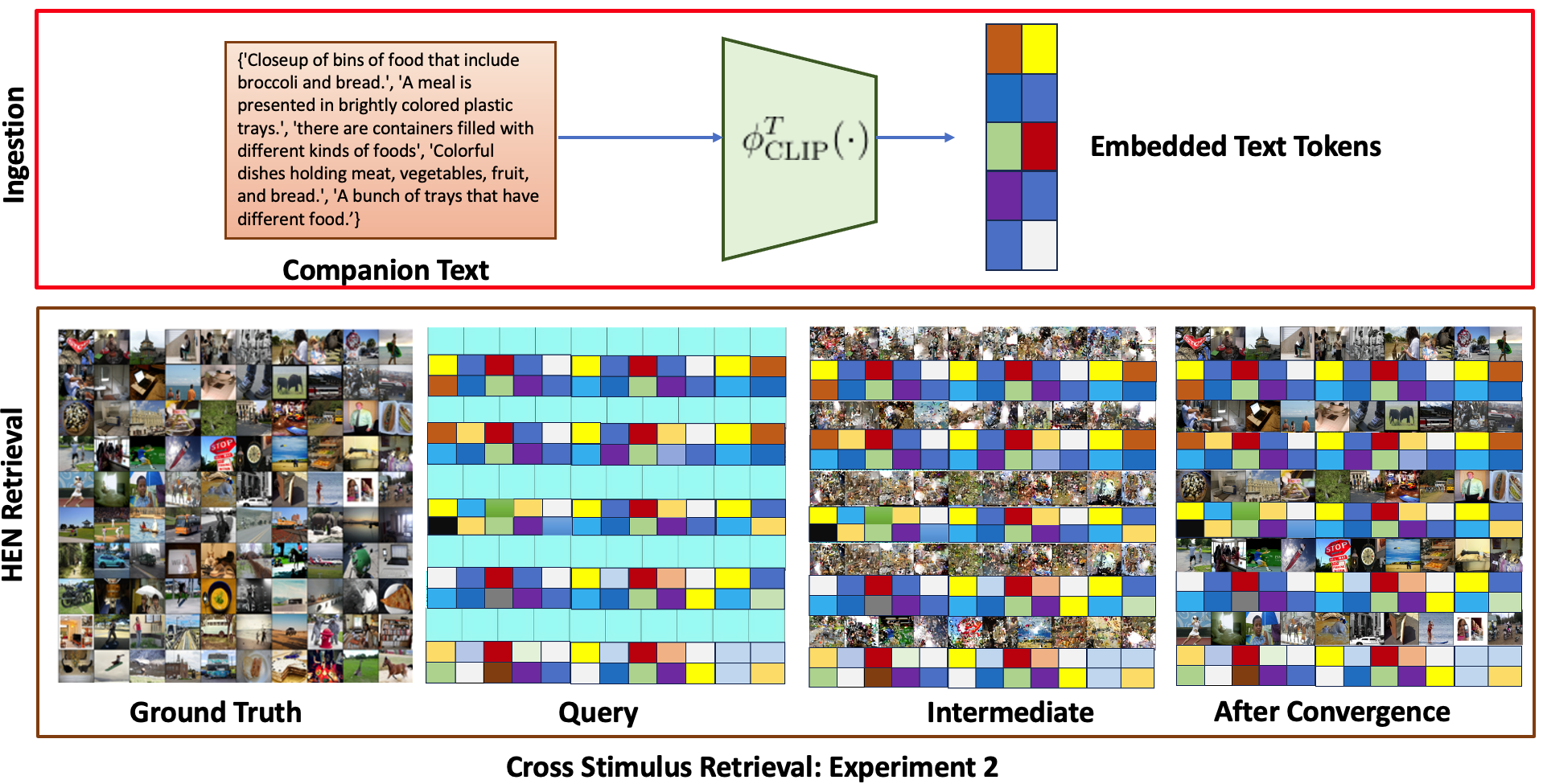}
\caption{Step-by-step progression of a cross-modal query using two different encodings for associated visual and language cues.  \textbf{Top Row:} The text stimulus is encoded via the CLIP~\citep{radford2021learning} text encoder and associated with the  image represented by a D-VAE encoded vector. \textbf{Bottom Row:} The reconstruction process for heteroassociation. \textbf{(L-R)} Ground Truth, Iteration $t=0$ starting with a blank canvas with the provided CLIP Encoded text inputs as query prompts,  an intermediate update, and the full reconstruction. This demonstrates the network's ability to accurately reconstruct the image from a text-only input from a completely different stimulus space as the image content.}
\label{fig:cross-modal-CLIP}
\end{figure*}
\begin{figure}[t]
\centering    \includegraphics[width=0.7\columnwidth]{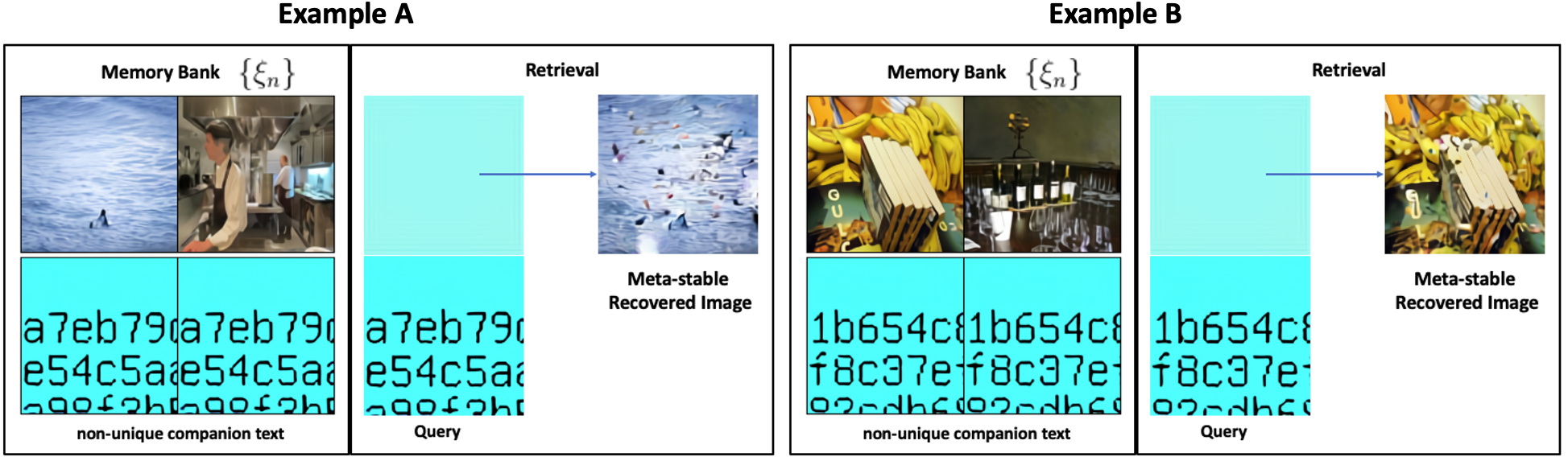}
\caption{In each example, \textbf{(Left)} the upper and lower image cases depict the disruption of unique associations in the $\{\hat{\mathbf{\xi}}_{n}\}$ memory bank. \textbf{(Right)} A single text input corresponding to these disrupted associations is used during the query phase. The top blue image represents an empty image as a zero-encoded vector at initialization. The reconstruction appears to be a meta-stable state. Neither of the original images is accurately recovered, supporting the hypothesis of unique text-image associations.}
    \label{fig:cm-break}
\end{figure}
We conducted three separate experiments. First, we use a native textual embedding for the language cue and associate it with the content to be stored as a practical way of enabling recall. Next, we explore the paradigm of stimuli type-conversion to render the language cue into a convenient image form to allow for content-based access. Finally, we show that if the uniqueness of association is lost, spurious memory states could again emerge even if the inputs are encoded. 

Fig.~\ref{fig:MHN} illustrates the overall methodology for the Hopfield Encoding Network (HEN) under hetero-associations. Here, the memory bank vectors are formed by concatenating image and text embeddings  $\hat{\mathbf{\xi}}_{n} = [\mathbf{\Phi^{I}_{\text{enc}}}(\mathbf{I}_{n}); \mathbf{\Phi^{T}_{\text{enc}}}(\mathbf{T}_{n})]$. During retrieval, we construct a query vector $\hat{\mathbf{s}}^{(0)} = [\mathbf{0};\hat{\mathbf{s}}_{\mathbf{T}}]$ constructed using the encoded text vector $\hat{\mathbf{s}}_{T} = \mathbf{\Phi^{T}_{\text{enc}}}(\mathbf{s}^{(0)}_{\mathbf{T}})$ and zeros in the location of the image encodings. Finally, after convergence $\hat{\mathbf{s}}^{(T_{f})} = [\hat{\mathbf{s}}^{(T_{f})}_{\mathbf{I}};\hat{\mathbf{s}}^{(T_{f})}_{\mathbf{T}}]$, we decode the image embedding $\mathbf{s}^{(T_{f})}_{\mathbf{T}} = \mathbf{\Phi^{I}}_{\text{dec}}(\hat{\mathbf{s}}_{\mathbf{I}}^{(T_{f})})$ to retrieve the image content.

To test language-image associations, we utilized the unique set of captions associated with each image in the COCO dataset. In our experiment, we allowed image and text to be encoded with different encoder decoder architectures. Specifically, we retained the best performing encoder (D-VAE) for image encoding (See Fig.~\ref{fig:cross-modal-CLIP}) but the textual associative stimulus was encoded using the CLIP foundational model \citep{radford2021learning}.

To create a more meaningful embedding, we concatenated the set of caption sentences per image into a single long sentence. This sentence was then encoded using the pre-trained CLIP model. The resulting text and image encodings were then ingested into HEN, as illustrated in Fig.~\ref{fig:cross-modal-CLIP} (Top). The experiment yielded promising results. The performance of the 6,000 images tested was on par with that of the discrete VAE in the same embedding space. {\em We found this to be significant, as it suggests that text and image encoders can operate in disjoint spaces while still achieving accurate reconstructions}, provided the Hopfield energy landscapes are appropriately normalized. Further, the top row of Table.~\ref{tab:cross-modal-results} shows robust performance across all CLIP combinations.

We also explored an alternate strategy for representing cross-stimulus cues in a single input space by pixelizing the text representations into a unique `image' representation (See example in Fig.~\ref{fig:cm-break}) instead of a text-encoder. Converting this text into an image allows us to re-use the image encoder-decoder for both image and text portions of the queries and memory bank latent-space transformations. To our surprise, this schema provided comparable recovery performance to using dedicated image and text encoding strategies in Table.~\ref{tab:cross-modal-results}. See Section 4 in the Appendix for details on the pixelization.

Finally, to examine the uniqueness of association hypothesis, we designed an experiment in which two different images to be stored in HEN were selected at random and associated with the same textual pattern. We rendered the text in a pixelized form and used the same encoding as for image to remove the effect of separate encodings for image and text in testing the uniqueness aspect. We queried the system using the pixelized text to observe the type of images that would be reconstructed. Fig.~\ref{fig:cm-break} indicates that violating the uniqueness constraint led to spurious recall where the reconstructed image appeared to be a mixture of two different images. Thus HEN can support cross-stimuli associations and the recall is accurate if the associative text pattern is distinct per image. 

\section{Conclusions}
In this paper, we unified and explored the diverse energy formulations used across associative memory models, highlighting their theoretical interconnections and the shared focus on pattern separability. This foundational analysis informed our development of key enhancements for Modern Hopfield Networks (MHNs), specifically by integrating pattern encoders and decoders to enhance separability and reduce metastable states. Additionally, we demonstrated how this approach supports cross-stimuli associations using different encodings, as long as the uniqueness of association is maintained suggesting promising adaptability for applications that require cross-modal recall, like multimedia search or multi-sensor data fusion. These advancements mark a step toward improving the practicality and performance of Modern Hopfield Networks for real-world retrieval and storage tasks. 

Additionally, HEN's architecture shows potential for other associative tasks, such as temporal sequences or visual variations (e.g., images of the same entity from different angles). Prior studies on associative memory \cite{gutfreund1988processing, chaudhry2023long, shriwas2019multi, millidge2022universal} suggest that with minor modifications, HEN could be extended to sequential or contextual retrieval, supporting broader applications in dynamic and context-rich environments. 

While HEN has shown consistent performance, we recognize that its convergence and stability depend on factors such as the update rule, encoder-decoder selection, and time step $T_{f}$. Our results indicate that after a set number of iterations ($T_{f} = 100$ in our case), retrieval dynamics reach a stable plateau; however, the optimal $T_{f}$ may vary by dataset. Thus, while our approach delays metastable states, further tuning may be necessary in different contexts.

Looking forward, future work could investigate HEN’s scalability with even larger datasets by implementing fine-tuning steps to optimize VAE parameters, potentially further reducing metastable behavior. This approach could enhance HEN’s memory capacity, enabling it to support practical storage and retrieval use cases at scale. These extensions position HEN as a versatile framework, adaptable to diverse associative memory tasks and scalable to meet the demands of large-scale, heterogeneous data environments.

\bibliography{icml_refs}

\begin{thebibliography}{31}
\providecommand{\natexlab}[1]{#1}
\providecommand{\url}[1]{\texttt{#1}}
\expandafter\ifx\csname urlstyle\endcsname\relax
  \providecommand{\doi}[1]{doi: #1}\else
  \providecommand{\doi}{doi: \begingroup \urlstyle{rm}\Url}\fi

\bibitem[Almeida et~al.(2007)Almeida, Idiart, and Lisman]{coldspringmemory}
Almeida, L.~D., Idiart, M., and Lisman, J.~E.
\newblock Memory retrieval time and memory capacity of the ca3 network: Role of
  gamma frequency oscillations.
\newblock \emph{Learning \& Memory}, 14:\penalty0 795, 11 2007.
\newblock ISSN 10720502.

\bibitem[Amari(1972)]{amari1972learning}
Amari, S.-I.
\newblock Learning patterns and pattern sequences by self-organizing nets of
  threshold elements.
\newblock \emph{IEEE Transactions on computers}, 100\penalty0 (11):\penalty0
  1197--1206, 1972.

\bibitem[Barra et~al.(2018)Barra, Beccaria, and Fachechi]{barra2018new}
Barra, A., Beccaria, M., and Fachechi, A.
\newblock A new mechanical approach to handle generalized hopfield neural
  networks.
\newblock \emph{Neural Networks}, 106:\penalty0 205--222, 2018.

\bibitem[Bernier et~al.(2017)]{dghopfield}
Bernier, B. et~al.
\newblock Dentate gyrus contributes to retrieval as well as encoding: Evidence
  from context fear conditioning, recall, and extinction.
\newblock \emph{J Neuroscience}, 37\penalty0 (26):\penalty0 6359–6371, 2017.

\bibitem[Bruck \& Roychowdhury(1990)Bruck and Roychowdhury]{bruck1990number}
Bruck, J. and Roychowdhury, V.~P.
\newblock On the number of spurious memories in the hopfield model (neural
  network).
\newblock \emph{IEEE Transactions on Information Theory}, 36\penalty0
  (2):\penalty0 393--397, 1990.

\bibitem[Burns et~al.(2022)Burns, Haga, and Fukai]{burns2022multiscale}
Burns, T.~F., Haga, T., and Fukai, T.
\newblock Multiscale and extended retrieval of associative memory structures in
  a cortical model of local-global inhibition balance.
\newblock \emph{Eneuro}, 9\penalty0 (3), 2022.

\bibitem[Chaudhry et~al.(2023)Chaudhry, Zavatone-Veth, Krotov, and
  Pehlevan]{chaudhry2023long}
Chaudhry, H.~T., Zavatone-Veth, J.~A., Krotov, D., and Pehlevan, C.
\newblock Long sequence hopfield memory.
\newblock \emph{arXiv preprint arXiv:2306.04532}, 2023.

\bibitem[Demircigil et~al.(2017)Demircigil, Heusel, L{\"o}we, Upgang, and
  Vermet]{demircigil2017model}
Demircigil, M., Heusel, J., L{\"o}we, M., Upgang, S., and Vermet, F.
\newblock On a model of associative memory with huge storage capacity.
\newblock \emph{Journal of Statistical Physics}, 168:\penalty0 288--299, 2017.

\bibitem[Gillett et~al.(2020)Gillett, Pereira, and
  Brunel]{gillett2020characteristics}
Gillett, M., Pereira, U., and Brunel, N.
\newblock Characteristics of sequential activity in networks with temporally
  asymmetric hebbian learning.
\newblock \emph{Proceedings of the National Academy of Sciences}, 117\penalty0
  (47):\penalty0 29948--29958, 2020.

\bibitem[Gutfreund \& Mezard(1988)Gutfreund and
  Mezard]{gutfreund1988processing}
Gutfreund, H. and Mezard, M.
\newblock Processing of temporal sequences in neural networks.
\newblock \emph{Physical Review Letters}, 61\penalty0 (2):\penalty0 235, 1988.

\bibitem[Hoover et~al.(2023)Hoover, Liang, Pham, Panda, Strobelt, Chau, Zaki,
  and Krotov]{hoover2023energy}
Hoover, B., Liang, Y., Pham, B., Panda, R., Strobelt, H., Chau, D.~H., Zaki,
  M.~J., and Krotov, D.
\newblock Energy transformer, 2023.

\bibitem[Hopfield(1982)]{hopfield1982neural}
Hopfield, J.~J.
\newblock Neural networks and physical systems with emergent collective
  computational abilities.
\newblock \emph{Proceedings of the national academy of sciences}, 79\penalty0
  (8):\penalty0 2554--2558, 1982.

\bibitem[Iatropoulos et~al.(2022)Iatropoulos, Brea, and
  Gerstner]{iatropoulos2022kernel}
Iatropoulos, G., Brea, J., and Gerstner, W.
\newblock Kernel memory networks: A unifying framework for memory modeling.
\newblock \emph{Advances in neural information processing systems},
  35:\penalty0 35326--35338, 2022.

\bibitem[Kang \& Toyoizumi(2023)Kang and Toyoizumi]{kang2023hopfield}
Kang, L. and Toyoizumi, T.
\newblock Hopfield-like network with complementary encodings of memories.
\newblock \emph{Physical Review E}, 108\penalty0 (5):\penalty0 054410, 2023.

\bibitem[Karuvally et~al.(2023)Karuvally, Sejnowski, and
  Siegelmann]{karuvally2023general}
Karuvally, A., Sejnowski, T., and Siegelmann, H.~T.
\newblock General sequential episodic memory model.
\newblock In \emph{International Conference on Machine Learning}, pp.\
  15900--15910. PMLR, 2023.

\bibitem[Krotov \& Hopfield(2020)Krotov and Hopfield]{krotov2020large}
Krotov, D. and Hopfield, J.
\newblock Large associative memory problem in neurobiology and machine
  learning.
\newblock \emph{arXiv preprint arXiv:2008.06996}, 2020.

\bibitem[Krotov \& Hopfield(2016)Krotov and Hopfield]{krotov2016dense}
Krotov, D. and Hopfield, J.~J.
\newblock Dense associative memory for pattern recognition.
\newblock \emph{Advances in neural information processing systems}, 29, 2016.

\bibitem[Lin et~al.(2015)Lin, Maire, Belongie, Bourdev, Girshick, Hays, Perona,
  Ramanan, Zitnick, and Dollár]{lin2015microsoft}
Lin, T.-Y., Maire, M., Belongie, S., Bourdev, L., Girshick, R., Hays, J.,
  Perona, P., Ramanan, D., Zitnick, C.~L., and Dollár, P.
\newblock Microsoft coco: Common objects in context, 2015.

\bibitem[Martins et~al.(2023)Martins, Niculae, and McNamee]{martins2023sparse}
Martins, A., Niculae, V., and McNamee, D.~C.
\newblock Sparse modern hopfield networks.
\newblock In \emph{Associative Memory {\&} Hopfield Networks in 2023}, 2023.
\newblock URL \url{https://openreview.net/forum?id=zwqlV7HoaT}.

\bibitem[Millidge et~al.(2022)Millidge, Salvatori, Song, Lukasiewicz, and
  Bogacz]{millidge2022universal}
Millidge, B., Salvatori, T., Song, Y., Lukasiewicz, T., and Bogacz, R.
\newblock Universal hopfield networks: A general framework for single-shot
  associative memory models.
\newblock In \emph{International Conference on Machine Learning}, pp.\
  15561--15583. PMLR, 2022.

\bibitem[Radford et~al.(2021)Radford, Kim, Hallacy, Ramesh, Goh, Agarwal,
  Sastry, Askell, Mishkin, Clark, et~al.]{radford2021learning}
Radford, A., Kim, J.~W., Hallacy, C., Ramesh, A., Goh, G., Agarwal, S., Sastry,
  G., Askell, A., Mishkin, P., Clark, J., et~al.
\newblock Learning transferable visual models from natural language
  supervision.
\newblock In \emph{International conference on machine learning}, pp.\
  8748--8763. PMLR, 2021.

\bibitem[Ramesh et~al.(2021)Ramesh, Pavlov, Goh, Gray, Voss, Radford, Chen, and
  Sutskever]{ramesh2021zero}
Ramesh, A., Pavlov, M., Goh, G., Gray, S., Voss, C., Radford, A., Chen, M., and
  Sutskever, I.
\newblock Zero-shot text-to-image generation.
\newblock In \emph{International Conference on Machine Learning}, pp.\
  8821--8831. PMLR, 2021.

\bibitem[Ramsauer et~al.(2020)Ramsauer, Sch{\"a}fl, Lehner, Seidl, Widrich,
  Adler, Gruber, Holzleitner, Pavlovi{\'c}, Sandve,
  et~al.]{ramsauer2020hopfield}
Ramsauer, H., Sch{\"a}fl, B., Lehner, J., Seidl, P., Widrich, M., Adler, T.,
  Gruber, L., Holzleitner, M., Pavlovi{\'c}, M., Sandve, G.~K., et~al.
\newblock Hopfield networks is all you need.
\newblock \emph{arXiv preprint arXiv:2008.02217}, 2020.

\bibitem[Rombach et~al.(2021)Rombach, Blattmann, Lorenz, Esser, and
  Ommer]{rombach2021high}
Rombach, R., Blattmann, A., Lorenz, D., Esser, P., and Ommer, B.
\newblock High-resolution image synthesis with latent diffusion models. 2022
  ieee.
\newblock In \emph{CVF Conference on Computer Vision and Pattern Recognition
  (CVPR)}, pp.\  10674--10685, 2021.

\bibitem[Shriwas et~al.(2019)Shriwas, Joshi, Ladwani, and
  Ramasubramanian]{shriwas2019multi}
Shriwas, R., Joshi, P., Ladwani, V.~M., and Ramasubramanian, V.
\newblock Multi-modal associative storage and retrieval using hopfield
  auto-associative memory network.
\newblock In \emph{International Conference on Artificial Neural Networks},
  pp.\  57--75. Springer, 2019.

\bibitem[Tyulmankov et~al.(2021)Tyulmankov, Fang, Vadaparty, and
  Yang]{tyulmankov2021biological}
Tyulmankov, D., Fang, C., Vadaparty, A., and Yang, G.~R.
\newblock Biological learning in key-value memory networks.
\newblock \emph{Advances in Neural Information Processing Systems},
  34:\penalty0 22247--22258, 2021.

\bibitem[Vaswani et~al.(2017)Vaswani, Shazeer, Parmar, Uszkoreit, Jones, Gomez,
  Kaiser, and Polosukhin]{vaswani2017attention}
Vaswani, A., Shazeer, N., Parmar, N., Uszkoreit, J., Jones, L., Gomez, A.~N.,
  Kaiser, {\L}., and Polosukhin, I.
\newblock Attention is all you need.
\newblock \emph{Advances in neural information processing systems}, 30, 2017.

\bibitem[Whittington et~al.(2020)Whittington, Muller, Mark, Chen, Barry,
  Burgess, and Behrens]{whittington2020tolman}
Whittington, J.~C., Muller, T.~H., Mark, S., Chen, G., Barry, C., Burgess, N.,
  and Behrens, T.~E.
\newblock The tolman-eichenbaum machine: unifying space and relational memory
  through generalization in the hippocampal formation.
\newblock \emph{Cell}, 183\penalty0 (5):\penalty0 1249--1263, 2020.

\bibitem[Widrich et~al.(2020)Widrich, Sch{\"a}fl, Pavlovi{\'c}, Ramsauer,
  Gruber, Holzleitner, Brandstetter, Sandve, Greiff, Hochreiter,
  et~al.]{widrich2020modern}
Widrich, M., Sch{\"a}fl, B., Pavlovi{\'c}, M., Ramsauer, H., Gruber, L.,
  Holzleitner, M., Brandstetter, J., Sandve, G.~K., Greiff, V., Hochreiter, S.,
  et~al.
\newblock Modern hopfield networks and attention for immune repertoire
  classification.
\newblock \emph{Advances in Neural Information Processing Systems},
  33:\penalty0 18832--18845, 2020.

\bibitem[Wright \& Gonzalez(2021)Wright and Gonzalez]{wright2021transformers}
Wright, M.~A. and Gonzalez, J.~E.
\newblock Transformers are deep infinite-dimensional non-mercer binary kernel
  machines.
\newblock \emph{arXiv preprint arXiv:2106.01506}, 2021.

\bibitem[Wu et~al.(2024)Wu, Hu, Hsiao, and Liu]{wu2024uniform}
Wu, D., Hu, J. Y.-C., Hsiao, T.-Y., and Liu, H.
\newblock Uniform memory retrieval with larger capacity for modern hopfield
  models.
\newblock \emph{arXiv preprint arXiv:2404.03827}, 2024.

\end{thebibliography}
\bibliographystyle{icml2024}

\section*{Appendix}
\appendix

\title{[Appendix] Modern Hopfield Networks meet Encoded Neural Representations - Addressing Practical Considerations}
\author{Satyananda Kashyap \\ IBM Research - Almaden 
\\ \texttt{satyananda.kashyap@ibm.com}  
\And Niharika S. D'Souza
\\ IBM Research - Almaden 
\And Luyao Shi 
\\ IBM Research - Almaden 
\And Ken C. L. Wong 
\\ IBM Research - Almaden 
\And Hongzhi Wang 
\\ IBM Research - Almaden 
\And Tanveer Syeda-Mahmood
\\ IBM Research - Almaden 
}





\vskip 0.3in
\section{Transformers and Kernel Memory Networks Representations}
From a representational perspective,~\citep{iatropoulos2022kernel} demonstrates that the formulations in Eqs.(1-2) in the main manuscript are special instances of Kernel Memory Networks (KMN), i.e. kernel machines from statistical learning, that train each individual neuron in the MHN to perform either kernel classification or interpolation (regression) with a minimum weight norm. Under this specialization, the update rule for an auto-associative memory framework (such as the classical Hopfield or Modern Hopfield Networks) can be expressed in a form analogous to Eq.~(2) in the main manuscript:
\begin{equation}
\mathbf{s}^{(t+1)} = \mathbf{\Xi} \mathbf{K}^{\dagger} {K}(\mathbf{\Xi}, \mathbf{s}^{(t)}),
\end{equation}
where the kernel function $K(\cdot,\cdot) : \mathcal{R}^{K} \times \mathcal{R}^{K} \rightarrow \mathcal{R}$ is defined pairwise between two input vectors in K-dimensional space, and $\mathbf{K} = K(\mathbf{\Xi}, \mathbf{\Xi}) = {\mathbf{\phi}(\mathbf{\Xi})}^{T} {\mathbf{\phi}(\mathbf{\Xi})}$ is the kernel matrix associated with the feature transformation $\mathbf{\phi}(\cdot): \mathcal{R}^{K} \rightarrow \mathcal{R}^{D}$ to a $D$ dimensional vector space. $\mathbf{K}^{\dagger}$ is its Moore-Penrose pseudoinverse, where $\mathbf{K}^{\dagger} = \mathbf{K}^{-1}$ if $\mathbf{\phi}({\mathbf\Xi})$ is full column rank. 

This framework is applicable under the restriction that $K(\cdot,\cdot)$ is a Mercer kernel (symmetric-positive definite), admits a reproducing bilinear form where the transformation $\mathbf{\Phi}(\cdot)$ is a reproducing kernel hilbert space (RKHS). Under these restrictions, ~\citep{iatropoulos2022kernel} lays the groundwork to analyze the storage capacity and corresponding recovery guarantees for query patterns (i.e. proximity to attractor basins) and their relation to the properties of the kernel function and the optimization procedure. In the case of continuous valued memories (non-binary case), a standard choice of a radial translation-invariant exponential kernel (with infinite dimensional basis) with a fixed spatial scale $r$ and temperature parameter $\alpha$.
\begin{equation}
K_{(\alpha,r)}(\mathbf{x},\mathbf{y}) = \exp{\Big[ - \Big(\frac{1}{r} {\vert\vert{\mathbf{x}-\mathbf{y}}\vert\vert}_{2}\Big)^{\alpha}\Big]}
\end{equation}
The KMN optimization is expected to be well-posed, convex, and is expected to provide global convergence to one of the stored memories (i.e. no meta-stable outputs) in theory.

\subsection{Relating Transformer and HEN Updates to the KMN formulation:}
Under the KMN formulation, the MHN form involves a kernel that is not symmetric positive definite, as the energy objective is not a convex optimization problem~\citep{iatropoulos2022kernel,wright2021transformers}. Nevertheless, it permits a bilinear reproducing form for the kernel $K(\cdot,\cdot)$, where the feature transformation maps $\{\mathbf{\Phi}_{\mathcal{X}}(\cdot): \mathcal{X} \rightarrow \mathcal{F}_{\mathcal{X}}(\cdot),\mathbf{\Phi}_{\mathcal{Y}}(\cdot) : \mathcal{Y} \rightarrow \mathcal{F}_{\mathcal{Y}}\}$ map input elements from vector spaces $\{\mathbf{x} \in \mathcal{X},\mathbf{y} \in \mathcal{Y}\}$ to the D-dimensional Banach spaces $\{{\mathbf{\Phi}_{\mathcal{X}}}(\cdot) \in \mathcal{F}_{\mathcal{X}},{\mathbf{\Phi}_{\mathcal{Y}}}(\cdot) \in \mathcal{F}_{\mathcal{Y}}\}$, characterized by linear manifold functional forms $\{\mathbf{f}_{\mathbf{\Phi}_{\mathcal{X}}}(\mathbf{x};\mathbf{W}_{Q}) \in \mathcal{B}_{\mathcal{X}}, \mathbf{g}_{\mathbf{\Phi}_{\mathcal{Y}}}(\mathbf{y};\mathbf{W}_{K}) \in \mathcal{B}_{\mathcal{Y}}\}$. The mathematical expression for the transformer kernel is:
\begin{equation*}
 K(\mathbf{x},\mathbf{y})= {\langle{\mathbf{\Phi}_{\mathcal{X}}(\mathbf{x}),\mathbf{\Phi}_{\mathcal{Y}}(\mathbf{y})}\rangle}_ {\mathcal{F}_{\mathcal{X}} \times \mathcal{F}_{\mathcal{Y}}} = {\langle{\mathbf{f}_{\mathbf{\Phi}_{\mathcal{X}}(x)},\mathbf{g}_{\mathbf{\Phi}_{\mathcal{Y}}(y)}}\rangle}_{\mathcal{B}_{\mathcal{X}} \times \mathcal{B}_{\mathcal{Y}}} = \exp{\Big[\frac{1}{\sqrt{d_{k}}} (\mathbf{W}_{Q}\mathbf{x})^T(\mathbf{W}_K\mathbf{y})\Big]}
\end{equation*}
From the representational perspective, it is sufficient to know the bilinear (reproducing) form of the kernel above in the MHN updates without requiring an explicit computation of \{$\mathbf{\Phi}_{\mathcal{X}(\mathbf{x})}, \mathbf{\Phi}_{\mathcal{Y}(\mathbf{y})}\}$ or sampling from the Banach spaces, $\mathcal{F}_{\mathcal{X}}: \mathbf{W}_{Q}\mathbf{x},\mathcal{F}_{\mathcal{Y}}: \mathbf{W}_{K}\mathbf{y}$, or estimation of the manifold functional forms $\{\mathcal{B}_{\mathcal{X}},\mathcal{B}_{\mathcal{Y}}\}$. Mathematically, these constructions can be formalized as: 
\begin{gather*}
\mathcal{B}_{\mathcal{X}} = \{\mathbf{f}_\mathbf{v}: \mathcal{X} \rightarrow \mathcal{R}: \mathbf{f}_{v}(\mathbf{x})=  \langle \mathbf{\Phi}_{\mathcal{X}(\mathbf{x})},\mathbf{v}\rangle_{\mathcal{F}_{X}\times \mathcal{F}_{Y}} ; \mathbf{v} \in \mathcal{F}_\mathcal{Y}, \mathbf{x} \in \mathcal{X} \} \\
= \Bigg\{ \mathbf{f}_{(\mathbf{W}_{K}\mathbf{y})}(\mathbf{x})= \exp{\Big[ \frac{(\mathbf{W}_{Q} \mathbf{x})^{T}\mathbf{W}_{K}\mathbf{y}}{\sqrt{d}_{k}} \Big]} ; \mathbf{W}_{K}\mathbf{y} \in \mathcal{F}_\mathcal{Y} , \mathbf{x} \in \mathcal{X} \Bigg\} \\
\mathcal{B}_{\mathcal{Y}} = \{\mathbf{g}_\mathbf{u}: \mathcal{Y} \rightarrow \mathcal{R}: \mathbf{g}_{u}(\mathbf{y})=  \langle {\mathbf{u},{\mathbf{\Phi}_{\mathcal{Y}(\mathbf{y})}}}\rangle_{\mathcal{F}_{X}\times \mathcal{F}_{Y}} ; \mathbf{u} \in \mathcal{F}_\mathcal{X}, \mathbf{y} \in \mathcal{Y} \} \\
= \Bigg\{ \mathbf{g}_{(\mathbf{W}_{Q}\mathbf{x})}(\mathbf{y})= \exp{\Big[ \frac{(\mathbf{W}_{Q} \mathbf{x})^{T}\mathbf{W}_{K}\mathbf{y}}{\sqrt{d}_{k}} \Big]} ; \mathbf{W}_{Q}\mathbf{x} \in \mathcal{F}_\mathcal{X} , \mathbf{y} \in \mathcal{Y} \Bigg\}
\end{gather*}
By extension, the kernel form for HEN can be formalized as follows:
\begin{gather*}
K_{\text{HEN}}(\mathbf{x},\mathbf{y}) =  \exp{\Big[ {\beta(\mathbf{\Phi}_{\textbf{enc}}(\mathbf{x}))^{T}(\mathbf{\Phi}_{\textbf{enc}}(\mathbf{y}))}\Big]} 
\\
\mathcal{B}_{\mathcal{X}} = 
\Bigg\{ \mathbf{f}_{(\mathbf{\Phi}_{\textbf{enc}}(\mathbf{y}))}(\mathbf{\Phi}_{\textbf{enc}}(\mathbf{x}))= \exp{\Big[ {{\beta} (\mathbf{\Phi}_{\textbf{enc}} (\mathbf{x}))^{T}\mathbf{\Phi}_{\textbf{enc}}(\mathbf{y})}\Big]} ; \mathbf{\Phi}_{\textbf{enc}}(\mathbf{y}) \in \mathcal{F}_\mathcal{Y} , \mathbf{x} \in \mathcal{X} \Bigg\} \\
\mathcal{B}_{\mathcal{Y}} = \Bigg\{ \mathbf{g}_{(\mathbf{\Phi}_{\textbf{enc}}(\mathbf{x}))}(\mathbf{\Phi}_{\textbf{enc}}(\mathbf{y}))= \exp{\Big[ {{\beta} (\mathbf{\Phi}_{\textbf{enc}} (\mathbf{x}))^{T}\mathbf{\Phi}_{\textbf{enc}}(\mathbf{y})}\Big]} ; \mathbf{\Phi}_{\textbf{enc}}(\mathbf{x}) \in \mathcal{F}_\mathcal{X} , \mathbf{y} \in \mathcal{Y} \Bigg\}
\end{gather*}

\subsection{Eq.~(2) is a special case of the Radial Translation-Invariant Exponential Kernel in Eq.~(4)}
According to the formulation in~\citep{iatropoulos2022kernel}, for continuous valued memories, the suggested form of kernel is a radial translation-invariant exponential kernel with a fixed spatial scale $r$ and parameter $\alpha$ is as follows:
\begin{gather*}
    K_{({\alpha,r})}(\mathbf{x},\mathbf{y}) = \exp{\Big[ - \Big(\frac{1}{r} {\vert\vert{\mathbf{x}-\mathbf{y}}\vert\vert}_{2}\Big)^{\alpha}\Big]} \\
     = \exp{\Big[ - \Big(\frac{1}{r} {\vert\vert{\mathbf{x}-\mathbf{y}}\vert\vert}^{2}_{2}\Big)^{\frac{\alpha}{2}}\Big]}  
     \\
     = \exp{\Big[ - \Big(\frac{1}{r} {{\vert\vert{\mathbf{x}}\vert\vert}^{2}_{2} + {\vert\vert{\mathbf{y}}\vert\vert}^{2}_{2} - 2  \mathbf{x}^{T} \mathbf{y}}\Big)^{\frac{\alpha}{2}}\Big]}  \\
     = \exp{\Big[ - \Big(\frac{1}{r} {[{\vert\vert{\mathbf{x}}\vert\vert}^{2}_{2} + {\vert\vert{\mathbf{y}}\vert\vert}^{2}_{2}]}\Big)^{\frac{\alpha}{2}}\Big]} \cdot \exp{\Big[ \Big(\frac{2}{r} \cdot {\frac{1}{2} \cdot {\mathbf{2} \mathbf{x}^{T}\mathbf{y}}}\Big)^{\frac{\alpha}{2}}\Big]} \\
     \text{Substituting} \ \  \alpha = 2, \ \ \beta = \frac{2}{r} = \frac{1}{\sqrt{d_{k}}} \ \ \text{and} \ \ \vert\vert{\mathbf{x}}\vert\vert_{2} = \vert\vert{\mathbf{y}}\vert\vert_{2} = 1 \\
     K_{\text{Trans}}(\mathbf{x},\mathbf{y}) = C \cdot \exp{\Big[\frac{1}{\sqrt{d_{k}}} (\mathcal{I}_{K}\mathbf{x})^{T}(\mathcal{I}_{K}\mathbf{y})\Big]} = C \cdot \exp{\Big[\beta \mathbf{x}^{T}\mathbf{y}\Big]}
\end{gather*}
where $C$ is a constant. This is the form of the transformer kernel corresponding to the update in Eq.~(2) and energy functional in Eq.~(1) of the main manuscript. We note that spherical normalization was also used in ~\citep{ramsauer2020hopfield,krotov2020large,wu2024uniform} as a mechanism to control the dynamics of the updates to mitigate metastable solutions.

\section{Ablations on $\beta$ for experiments on choices of $f_{\text{sim}}(\cdot,\cdot)$ and Neural Representation}
\label{sec:beta-ablation}

\begin{table}[h!]
    \caption{Structural Similarity Index Measure for HEN for increasing $\beta$ values. This table presents the 1-SSIM results (lower is better) for a set of N=6000 images from the MS-COCO dataset. Every row represents the similarity metrics (Dot or $\ell_2$) for five different types of pre-trained encoder-decoder architectures (kl8, vqf8, klf16, vqf16, dVAE). The Image column represents the Modern Hopfield Network formulation, the established baseline model in this comparison. We see that as for increasing temperature values of $\beta$, the recovery stabilizes and produces near exact recoveries for all sets of the trained encoder-decoder architectures. In comparison, we see that the image-based Modern Hopfield network fails to handle a large number of images in memory.}
    \resizebox{\columnwidth}{!}{%
    \begin{tabular}{|l||l|l|l|l|l|l||l|l|l|l|l|l|}
    \hline
        $\beta$ & \textbf{Dot-klf8} & \textbf{Dot-vqf8} & \textbf{Dot-klf16} & \textbf{Dot-vqf16} & \textbf{Dot-Image} & \textbf{Dot-dVAE} & \textbf{L2-klf8} & \textbf{L2-vqf8} & \textbf{L2-klf16} & \textbf{L2-vqf16} & \textbf{L2-Image} & \textbf{L2-dVAE} \\ \hline
        500 & 0.0003 & 0.0213 & 0.0043 & 0.0191 & 0.9089 & 0.0000 & 0.1840 & 0.0523 & 0.2315 & 0.0819 & 0.8815 & 0.0000 \\ \hline
        480 & 0.0003 & 0.0213 & 0.0043 & 0.0191 & 0.9089 & 0.0000 & 0.1840 & 0.0523 & 0.2315 & 0.0819 & 0.8815 & 0.0000 \\ \hline
        460 & 0.0003 & 0.0213 & 0.0043 & 0.0191 & 0.9089 & 0.0000 & 0.1840 & 0.0523 & 0.2315 & 0.0819 & 0.8815 & 0.0000 \\ \hline
        440 & 0.0003 & 0.0213 & 0.0043 & 0.0191 & 0.9089 & 0.0000 & 0.1840 & 0.0523 & 0.2315 & 0.0819 & 0.8815 & 0.0000 \\ \hline
        420 & 0.0003 & 0.0213 & 0.0043 & 0.0191 & 0.9089 & 0.0000 & 0.1840 & 0.0523 & 0.2315 & 0.0819 & 0.8815 & 0.0000 \\ \hline
        400 & 0.0003 & 0.0213 & 0.0043 & 0.0191 & 0.9089 & 0.0000 & 0.1840 & 0.0523 & 0.2315 & 0.0819 & 0.8815 & 0.0000 \\ \hline
        380 & 0.0003 & 0.0213 & 0.0043 & 0.0191 & 0.9089 & 0.0000 & 0.1840 & 0.0523 & 0.2315 & 0.0819 & 0.8815 & 0.0000 \\ \hline
        360 & 0.0003 & 0.0213 & 0.0043 & 0.0191 & 0.9089 & 0.0000 & 0.1840 & 0.0523 & 0.2315 & 0.0819 & 0.8815 & 0.0000 \\ \hline
        340 & 0.0003 & 0.0213 & 0.0043 & 0.0191 & 0.9089 & 0.0000 & 0.1840 & 0.0523 & 0.2315 & 0.0819 & 0.8815 & 0.0000 \\ \hline
        320 & 0.0003 & 0.0213 & 0.0043 & 0.0191 & 0.9089 & 0.0000 & 0.1840 & 0.0523 & 0.2315 & 0.0819 & 0.8815 & 0.0000 \\ \hline
        300 & 0.0003 & 0.0213 & 0.0043 & 0.0191 & 0.9089 & 0.0000 & 0.1840 & 0.0523 & 0.2315 & 0.0819 & 0.8815 & 0.0000 \\ \hline
        280 & 0.0003 & 0.0213 & 0.0043 & 0.0191 & 0.9089 & 0.0000 & 0.1840 & 0.0523 & 0.2315 & 0.0819 & 0.8815 & 0.0000 \\ \hline
        260 & 0.0003 & 0.0213 & 0.0043 & 0.0191 & 0.9089 & 0.0000 & 0.1840 & 0.0523 & 0.2315 & 0.0819 & 0.8815 & 0.0000 \\ \hline
        240 & 0.0003 & 0.0213 & 0.0043 & 0.0191 & 0.9090 & 0.0000 & 0.1840 & 0.0523 & 0.2315 & 0.0819 & 0.8815 & 0.0000 \\ \hline
        220 & 0.0003 & 0.0213 & 0.0043 & 0.0191 & 0.9090 & 0.0000 & 0.1840 & 0.0523 & 0.2315 & 0.0819 & 0.8815 & 0.0000 \\ \hline
        200 & 0.0003 & 0.0213 & 0.0043 & 0.0191 & 0.9499 & 0.0000 & 0.1840 & 0.0523 & 0.2315 & 0.0819 & 0.8815 & 0.0000 \\ \hline
        180 & 0.0003 & 0.0213 & 0.0043 & 0.0191 & 0.9503 & 0.0000 & 0.1840 & 0.0523 & 0.2315 & 0.0819 & 0.8815 & 0.0000 \\ \hline
        160 & 0.0003 & 0.0213 & 0.0043 & 0.0191 & 0.9506 & 0.0000 & 0.1840 & 0.0523 & 0.2315 & 0.0819 & 0.8815 & 0.0000 \\ \hline
        140 & 0.0003 & 0.0213 & 0.0043 & 0.0191 & 0.9514 & 0.0000 & 0.1840 & 0.0523 & 0.2315 & 0.0819 & 0.8815 & 0.0000 \\ \hline
        120 & 0.0003 & 0.0213 & 0.0043 & 0.0191 & 0.9521 & 0.0000 & 0.1840 & 0.0523 & 0.2315 & 0.0819 & 0.8815 & 0.0000 \\ \hline
        100 & 0.0003 & 0.0213 & 0.6657 & 0.0191 & 0.9518 & 0.0000 & 0.1840 & 0.0523 & 0.2315 & 0.0819 & 0.8815 & 0.0000 \\ \hline
        80 & 0.0003 & 0.0213 & 0.6625 & 0.0191 & 0.9517 & 0.0000 & 0.1840 & 0.0523 & 0.2315 & 0.0819 & 0.8815 & 0.0000 \\ \hline
        60 & 0.0003 & 0.0213 & 0.6617 & 0.0191 & 0.9501 & 0.7911 & 0.1840 & 0.0523 & 0.2315 & 0.0819 & 0.8815 & 0.0000 \\ \hline
        40 & 0.0002 & 0.0213 & 0.6644 & 0.0191 & 0.9481 & 0.7629 & 0.1840 & 0.0523 & 0.2315 & 0.0819 & 0.8815 & 0.0000 \\ \hline
        20 & 0.6635 & 0.0213 & 0.6665 & 0.7104 & 0.9464 & 0.7828 & 0.1840 & 0.0523 & 0.2315 & 0.0819 & 0.8815 & 0.0000 \\ \hline
    \end{tabular}}
    \label{tab:ssim}
\end{table}
\begin{table}[h!]
\caption{Mean Squared Error (MSE) Measure for HEN for increasing $\beta$ values. This table presents the MSE results (lower is better) for a set of N=6000 images from the MS-COCO dataset. Every row represents the similarity metrics (Dot or $\ell_2$) for five different types of pre-trained encoder-decoder architectures (kl8, vqf8, klf16, vqf16, dVAE). The Image column represents the Modern Hopfield Network formulation, the established baseline model in this comparison. We see that as for increasing temperature values of $\beta$, the recovery stabilizes and produces near exact recoveries for all sets of the trained encoder-decoder architectures. In comparison, we see that the image-based Modern Hopfield network fails to handle a large number of images in memory.}
    \resizebox{\columnwidth}{!}{%
    \begin{tabular}{|l|l|l|l|l|l|l|l|l|l|l|l|l|}
    \hline
        $\beta$ & \textbf{Dot-klf8} & \textbf{Dot-vqf8} & \textbf{Dot-klf16} & \textbf{Dot-vqf16} & \textbf{Dot-Image} & \textbf{Dot-dVAE} & \textbf{L2-klf8} & \textbf{L2-vqf8} & \textbf{L2-klf16} & \textbf{L2-vqf16} & \textbf{L2-Image} & \textbf{L2-dVAE} \\ \hline
        500 & 0.0000 & 0.0005 & 0.0001 & 0.0003 & 0.1788 & 0.0000 & 0.0280 & 0.0057 & 0.0341 & 0.0105 & 0.0906 & 0.0000 \\ \hline
        480 & 0.0000 & 0.0005 & 0.0001 & 0.0003 & 0.1788 & 0.0000 & 0.0280 & 0.0057 & 0.0341 & 0.0105 & 0.0906 & 0.0000 \\ \hline
        460 & 0.0000 & 0.0005 & 0.0001 & 0.0003 & 0.1788 & 0.0000 & 0.0280 & 0.0057 & 0.0341 & 0.0105 & 0.0906 & 0.0000 \\ \hline
        440 & 0.0000 & 0.0005 & 0.0001 & 0.0003 & 0.1788 & 0.0000 & 0.0280 & 0.0057 & 0.0341 & 0.0105 & 0.0906 & 0.0000 \\ \hline
        420 & 0.0000 & 0.0005 & 0.0001 & 0.0003 & 0.1788 & 0.0000 & 0.0280 & 0.0057 & 0.0341 & 0.0105 & 0.0906 & 0.0000 \\ \hline
        400 & 0.0000 & 0.0005 & 0.0001 & 0.0003 & 0.1788 & 0.0000 & 0.0280 & 0.0057 & 0.0341 & 0.0105 & 0.0906 & 0.0000 \\ \hline
        380 & 0.0000 & 0.0005 & 0.0001 & 0.0003 & 0.1788 & 0.0000 & 0.0280 & 0.0057 & 0.0341 & 0.0105 & 0.0906 & 0.0000 \\ \hline
        360 & 0.0000 & 0.0005 & 0.0001 & 0.0003 & 0.1788 & 0.0000 & 0.0280 & 0.0057 & 0.0341 & 0.0105 & 0.0906 & 0.0000 \\ \hline
        340 & 0.0000 & 0.0005 & 0.0001 & 0.0003 & 0.1788 & 0.0000 & 0.0280 & 0.0057 & 0.0341 & 0.0105 & 0.0906 & 0.0000 \\ \hline
        320 & 0.0000 & 0.0005 & 0.0001 & 0.0003 & 0.1788 & 0.0000 & 0.0280 & 0.0057 & 0.0341 & 0.0105 & 0.0906 & 0.0000 \\ \hline
        300 & 0.0000 & 0.0005 & 0.0001 & 0.0003 & 0.1788 & 0.0000 & 0.0280 & 0.0057 & 0.0341 & 0.0105 & 0.0906 & 0.0000 \\ \hline
        280 & 0.0000 & 0.0005 & 0.0001 & 0.0003 & 0.1787 & 0.0000 & 0.0280 & 0.0057 & 0.0341 & 0.0105 & 0.0906 & 0.0000 \\ \hline
        260 & 0.0000 & 0.0005 & 0.0001 & 0.0003 & 0.1787 & 0.0000 & 0.0280 & 0.0057 & 0.0341 & 0.0105 & 0.0906 & 0.0000 \\ \hline
        240 & 0.0000 & 0.0005 & 0.0001 & 0.0003 & 0.1787 & 0.0000 & 0.0280 & 0.0057 & 0.0341 & 0.0105 & 0.0906 & 0.0000 \\ \hline
        220 & 0.0000 & 0.0005 & 0.0001 & 0.0003 & 0.1786 & 0.0000 & 0.0280 & 0.0057 & 0.0341 & 0.0105 & 0.0906 & 0.0000 \\ \hline
        200 & 0.0000 & 0.0005 & 0.0001 & 0.0003 & 0.0913 & 0.0000 & 0.0280 & 0.0057 & 0.0341 & 0.0105 & 0.0906 & 0.0000 \\ \hline
        180 & 0.0000 & 0.0005 & 0.0001 & 0.0003 & 0.0899 & 0.0000 & 0.0280 & 0.0057 & 0.0341 & 0.0105 & 0.0906 & 0.0000 \\ \hline
        160 & 0.0000 & 0.0005 & 0.0001 & 0.0003 & 0.0833 & 0.0000 & 0.0280 & 0.0057 & 0.0341 & 0.0105 & 0.0906 & 0.0000 \\ \hline
        140 & 0.0000 & 0.0005 & 0.0001 & 0.0003 & 0.0824 & 0.0000 & 0.0280 & 0.0057 & 0.0341 & 0.0105 & 0.0906 & 0.0000 \\ \hline
        120 & 0.0000 & 0.0005 & 0.0001 & 0.0003 & 0.0815 & 0.0000 & 0.0280 & 0.0057 & 0.0341 & 0.0105 & 0.0906 & 0.0000 \\ \hline
        100 & 0.0000 & 0.0005 & 0.1112 & 0.0003 & 0.0758 & 0.0000 & 0.0280 & 0.0057 & 0.0341 & 0.0105 & 0.0906 & 0.0000 \\ \hline
        80 & 0.0000 & 0.0005 & 0.1082 & 0.0003 & 0.0746 & 0.0000 & 0.0280 & 0.0057 & 0.0341 & 0.0105 & 0.0906 & 0.0000 \\ \hline
        60 & 0.0000 & 0.0005 & 0.1074 & 0.0003 & 0.0698 & 0.4034 & 0.0280 & 0.0057 & 0.0341 & 0.0105 & 0.0906 & 0.0000 \\ \hline
        40 & 0.0000 & 0.0005 & 0.1070 & 0.0003 & 0.0659 & 0.3819 & 0.0280 & 0.0057 & 0.0341 & 0.0105 & 0.0906 & 0.0000 \\ \hline
        20 & 0.1434 & 0.0005 & 0.0852 & 0.1447 & 0.0635 & 0.3821 & 0.0280 & 0.0057 & 0.0341 & 0.0105 & 0.0906 & 0.0000 \\ \hline
    \end{tabular}}
    \label{tab:mse}
\end{table}

In this experiment, we examine the impact of the $\beta$ parameter on the performance of HEN in the context of image recovery, the choice of the similarity function $f_{\text{sim}}(\cdot,\cdot)$ in the energy functional. We explore the influence of various pre-trained neural encoder-decoder architectures on the HEN's ability to converge to stable states and achieve accurate recovery. As in the main manuscript, we evaluated the performance using the structural similarity index measure (1-SSIM) and the Mean squared error (MSE) to assess the image recovery. The experiment was conducted using the MS-COCO dataset subset of $6000$ images. This dataset was selected for its diversity and complexity, providing a robust challenge to the storage and recovery capabilities of the Modern Hopfield networks and Hopfield Encoded Networks. 

The $\beta$ value was varied from $20-500$ in increments of $20$. This range was chosen to cover a broad range of the dynamics within the MHN. For the similarity function $f_{\text{sim}}(\cdot,\cdot)$ , we examine the $\ell_{2}$ distance and the dot product. The choices for pre-trained encoder-decoder architectures are the Discrete Variational Autoencoder (D-VAE) and variants of VQ-K8 and VQ-F16. We also compared the performance of the Hopfield encoded networks with that of image-based Modern Hopfield networks that did not utilize any pre-trained encoder-decoder systems. This comparison sheds light on the benefits of incorporating complex encoding mechanisms into Hopfield networks. To maintain consistency in representation, we downsampled the images to a resolution of 28 × 28 × 3. 

Tables \ref{tab:ssim}, \ref{tab:mse} highlight the critical role of \(\beta\) in tuning the network's dynamics, with optimal performance observed beyond specific \(\beta\) thresholds. The use of encoded neural representations is stable across a much larger range of $\beta$s. In conjunction with the results on separability of encodings and the results in Section~2.4.1 of main manuscript, this experiment highlights the benefits of leveraging encoded neural representations for addressing practical storage considerations in the MHN framework over alternatives such as kernel based approaches. 

In exploring the intricate dynamics of HEN, our investigation does not extend to a universal assertion that any combination of time step $T_{f}$, update rule/energy function, and well-trained encoder-decoder methods will provide good convergence for all datasets or patterns. The behavior of the update dynamics tends to stabilize after a certain number of iterations ($T_{f}$), beyond which further improvements are marginal. In our experiments, we found that after $T_{f} = 100$ iterations, the updates plateau, suggesting that the optimal value of $T_{f}$ is both dataset-dependent and influenced by factors like data dimensionality and pattern separability.

Additionally, delaying the onset of metastable states is an important feature of our proposed approach, as it enables the system to store more content within encoded representations, potentially leading to a higher memory-capacity ceiling. However, this does not imply that the method is entirely free from metastable states or that it will always guarantee convergence for all possible input distributions. 

Certain real-world datasets with high noise, irregular structure, or non-separable patterns may still present challenges, even for our enhanced architecture. Future work could involve extending the analysis of metastable state dynamics across different datasets and exploring further variations of update rules or encoding strategies to improve convergence guarantees across a wider range of practical applications.

\section{Quantifying Meta-Stable States}
Recall that SSIM and MSE metrics used in Fig.~2 in the main manuscript quantify the error in the recovery of the stored image content (Fig.~1, column~4). By definition, meta-stable states correspond to degenerate solutions (among entities in the recovered content) in the energy landscape. To quantify the prevalence of these meta-stable states as a function of the dynamics of the HEN, we report an additional proxy for identifying such behavior. 

\begin{figure*}[]
\centering
\fbox{\includegraphics[scale=0.4]{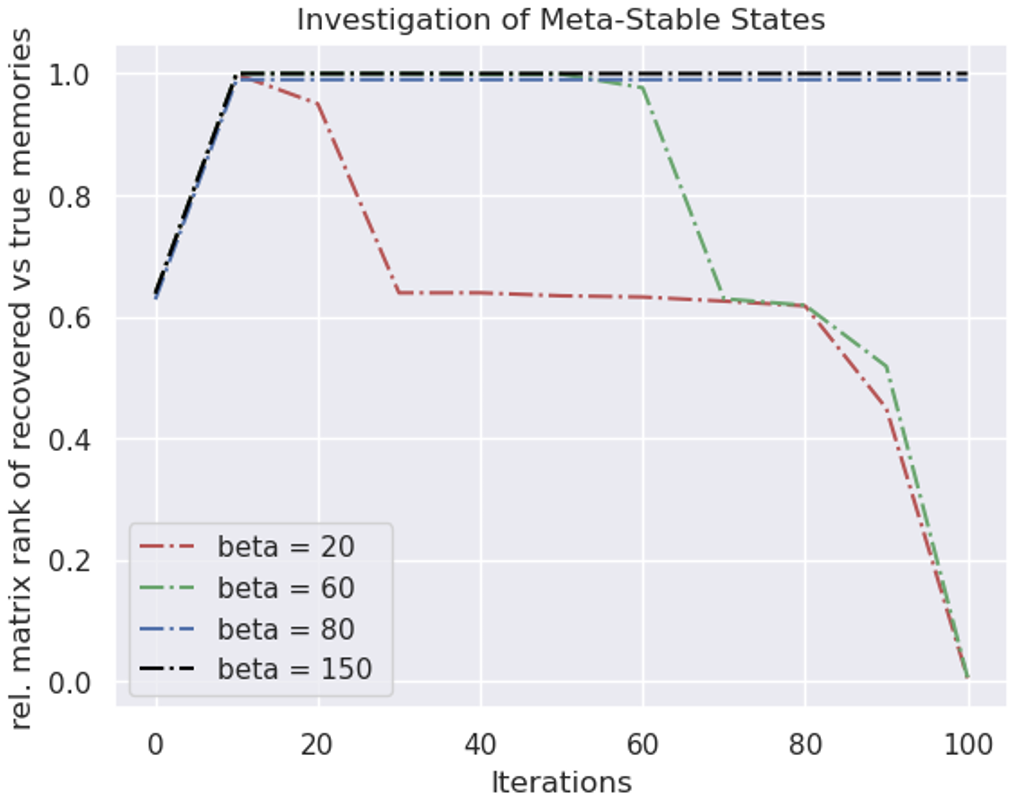}}
\caption{Tracking the evolution of meta-stable states during the dynamics of the HEN. We plot the relative rank ($ \text{RR} = R_{\hat{\mathbf{S}}}/R_{\hat{\mathbf{\Xi}}}$) of the recovered state matrix ($\hat{\mathbf{S}} = [\hat{\mathbf{s}}_{1}, \dots ,\hat{\mathbf{s}}_{N}]$ ) to that of the memory bank ($\hat{\mathbf{\Xi}} = [\hat{\mathbf{\xi}}_{1}, \dots ,\hat{\mathbf{\xi}}_{N}]$). Lower matrix rank signals the presence of meta-stable states. For a judiciously chosen $\beta$, the HEN provides near-perfect recall for this collection of images}
\label{MatRank}
\end{figure*}

Specifically, we report the relative rank ($ \text{RR} = R_{\hat{\mathbf{S}}}/R_{\hat{\mathbf{\Xi}}}$) of the recovered state matrix ($\hat{\mathbf{S}} = [\hat{\mathbf{s}}_{1}, \dots ,\hat{\mathbf{s}}_{N}]$ ) to that of the memory bank ($\hat{\mathbf{\Xi}} = [\hat{\mathbf{\xi}}_{1}, \dots ,\hat{\mathbf{\xi}}_{N}]$) for various values of $\beta$ during the evolution of the dynamics in Eq.~(2). In Fig.~\ref{MatRank}, we plot the RR on the y axis against the number of iterations on the x axis. 

$RR<1$ upon convergence indicates degeneracy in the recovered solutions. Each dashed line corresponds to different values of the temperature $\beta$. For this illustration, we use the dVAE encoder with the dot product formulation of the HEN and 6000 images from the MS-COCO dataset (i.e. the same setup as the green line in Fig.~2 of the main manuscript). We observe very consistent trends with those seen in the 1-SSIM and MSE metrics, where near-zero values of these metrics signal no degeneracy (no meta-stable states), while increasing non-zero values indicate imperfect retrieval due to the presence of meta-stable states (such as with row 1, column 4 of Fig.~2). Early in the optimization ($t<20$ iterations), the $\{\hat{\mathbf{s}}^{(t)}_{i}\}$ contain information corresponding to encoding the masked image signal. These vectors are distinct across input examples and also exhibit some baseline (non-zero) correlation with the stored memories $\{\hat{\mathbf{\xi}}_{i}\}$ . As the dynamics evolve, we observe that for low values of $\beta =[20,50]$, the dynamics destabilize to low-rank solutions, collapsing the retrieval fidelity. For sufficiently high $\beta =[80,150]$, the dynamics stabilize over a period of time, leading to near-perfect recovery ( consistent with $1-\text{SSIM}=\text{MSE}=0$)

\textcolor{black}{Finally, the work of \citep{wu2024uniform} leverages the equivalence between MHNs, kernel methods, and transformers to directly address the issue of separability of patterns under arbitrary data-distributions. Their framework introduces a second separation maximization objective that is added to the energy minimization objective (i.e. Eq.(1) in the main manuscript). The separation objective is based on a kernel transformation parameterized as a learnable linear kernel or deep network (i.e. encoder) and radial kernel. Thus, pattern retrieval is a two step process that interleaves a stochastic optimization of the encoding parameters with the retrieval dynamics of the Modern Hopfield Network. Here, in each step, local convergence is achieved either as a mini-batch or full batch gradient descent routine. This strategy was shown to provide improvements over the traditional MHNs and their variants upto dataset sizes of 500 for image retrieval. In contrast, our framework fixes the encoding transformation as defined by a pre-trained VAE. As the VAE parameters are frozen apriori, this avoids an expensive outer optimization loop, in turn providing expedient retrieval for datasets of relatively larger size (15000 examples) as well as support hetero-association.} 

\section{Handling cross-stimuli using pixelized language-image association}
\label{pixelized}
To test language-image associations, we utilized the unique set of captions associated with each image in the COCO dataset. Specifically,  we employed Python's \texttt{hashlib.sha256(.)} function to hash the captions generating a unique ID text string to associate with the image. Initially, we created a memory bank~$\{{\hat{\mathbf{\xi}}}^{\bf T}_{n} \}$ by converting the hashed captions into pixelized text representations using a generic text-to-pixel function. Subsequently, both the pixelized text and the corresponding images were processed through the same encoder. The resulting vectors were concatenated to form the elements of the memory bank $\{\hat{\mathbf{\xi}}_{n}  = [\hat{\mathbf{\xi}}^{\bf I}_{n} ; {\hat{\mathbf{\xi}}}^{\bf T}_{n}] \}$.

\begin{figure*}[ht!]
\centering
\includegraphics[width=\textwidth]{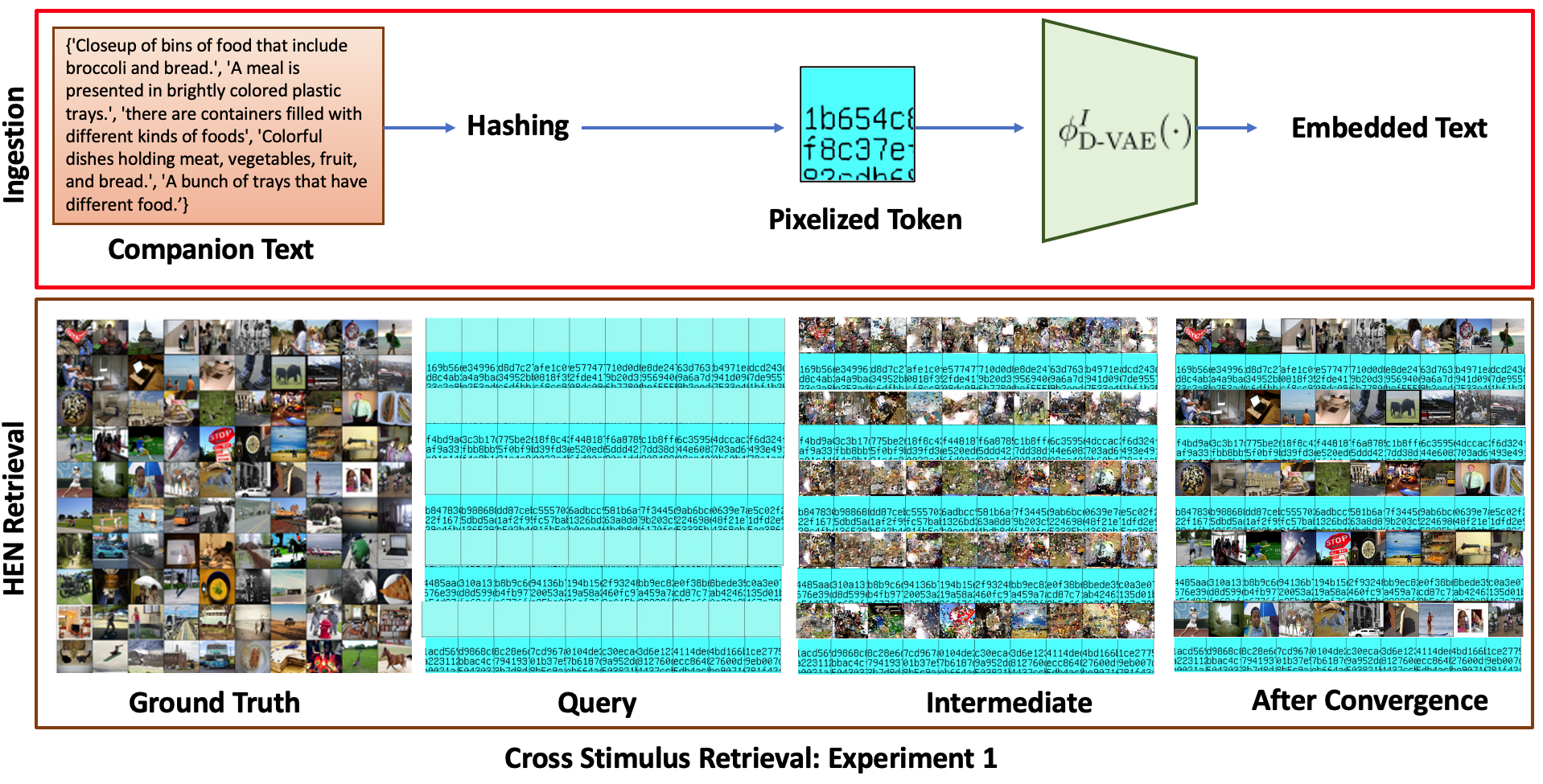}
\caption{Step-by-step progression of a cross-modal query using only text input. \textbf{Top Row:} The companion text is pixelized and encoded (using D-VAE in this example). This encoded representation is used to reconstruct the complete image. \textbf{Bottom Row:} The reconstruction process for the experiment in Subsection~\ref{pixelized}. \textbf{(L-R)} Ground Truth, Iteration $t=0$ starting with a blank canvas with the provided pixelized text inputs as query prompts,  an intermediate update, and finally the fully reconstructed image. This visualization effectively demonstrates the network's ability to accurately reconstruct the image from a text-only input modified into an image-based representation }
    \label{fig:cross-modal-pix}
\end{figure*}

During the query phase, we supplied only the pixelized text part of the encoded vector, setting the image component to zero. The Hopfield network iteratively updated the image encoding vector, which was passed through the corresponding decoder to reconstruct the image. Fig.~\ref{fig:cross-modal-pix} illustrates the network's progression in reconstructing the image based on the pixelized text input. The recurrent updates in the HEN iteratively reconstructed the full image.

As illustrated in Figs.~\ref{fig:cross-modal-pix}, the HEN network is able to recall perfectly using pixelized cross-stimuli associations. 
Table~\ref{tab:cross-modal-results} (repeated from the main section of the paper for convenience) reveals that all HEN variants with different encodings still outperformed traditional image-based MHNs even as the number of image patterns to store increased. While the HEN can recall accurately based on cross-stimuli associations, we expect such associations to be unique as in the case of stimuli from the same domain/modality (See result in Fig.~6 of the main manuscript). 

\begin{table*}[]
\caption{The table displays the performance of various encoded cross-modal HEN compared to image-based Modern Hopfield networks as the memory bank $\{\hat{\mathbf{\xi}}_{n}\}$ increases. All of the different encoders performed well. The first line shows the CLIP-encoded cross-modal representations, while the following lines present pixelized text-encoded representations for increasing image sizes.}
\label{tab:my-table}
\centering
\medskip
\resizebox{0.7\columnwidth}{!}{%
\begin{tabular}{@{}ccccc@{}}
\toprule
\multicolumn{1}{l}{\begin{tabular}[c]{@{}l@{}}NUM IMAGES\\ 1-SSIM, MSE\end{tabular}} & \multicolumn{1}{l}{vqf8} & \multicolumn{1}{l}{vqf16} & \multicolumn{1}{l}{Image} & \multicolumn{1}{l}{D-VAE} \\ \midrule
6000-CLIP & 0.016, 0.000 & 0.024, 0.000 & 0.681, 0.118 & \textbf{0.000, 0.000} \\ \midrule
6000 & 0.016, 0.000 & 0.024, 0.000 & 0.952, 0.214 & \textbf{0.000, 0.000} \\ \midrule
8000 & 0.016, 0.000 & 0.023, 0.000 & 0.952, 0.215 & \textbf{0.000, 0.000} \\ \midrule
10000 & 0.015, 0.000 & 0.024, 0.000 & 0.952, 0.215 & \textbf{0.000, 0.000} \\ \midrule
15000 & 0.015, 0.000 & 0.024, 0.000 & 0.952, 0.215 & \textbf{0.000, 0.000} \\ \bottomrule
\end{tabular}%
\label{tab:cross-modal-results}
}
\end{table*}


\end{document}